\documentclass[journal]{IEEEtran}
\usepackage{cite}
\usepackage{amsmath,amssymb,amsfonts}
\usepackage{algorithmic}
\usepackage{graphicx}
\usepackage{textcomp}
\usepackage{subcaption}
\usepackage{dblfloatfix}
\usepackage[utf8]{inputenc}
\usepackage{hyperref}
\usepackage{caption}

\newcommand{\orcidicon}[1]{%
  \href{https://orcid.org/#1}{\includegraphics[height=0.62em]{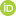}}%
}

\title{PerfCam: Digital Twinning for Production Lines Using 3D Gaussian Splatting and Vision Models}

\author{Michel Gokan Khan\orcidicon{0000-0002-4825-8831},~\IEEEmembership{Member,~IEEE},
Renan Guarese\orcidicon{0000-0003-1206-5701},~\IEEEmembership{Member,~IEEE},
Fabian Johnson,
Xi Vincent Wang\orcidicon{0000-0001-9694-0483},
Anders Bergman,
Benjamin Edvinsson,
Mario Romero\orcidicon{0000-0003-4616-189X},~\IEEEmembership{Member,~IEEE},
Jérémy Vachier\orcidicon{0000-0002-9043-4034},
and Jan Kronqvist\orcidicon{0000-0003-0299-5745}%
\thanks{Corresponding author: Michel Gokan Khan (e-mail: michelgk@kth.se).}%
\thanks{M.~Gokan~Khan and J.~Kronqvist are with the School of Engineering Sciences, KTH Royal Institute of Technology, Stockholm, Sweden (e‑mail: \{michelgk, jankr\}@kth.se).}%
\thanks{M.~Gokan~Khan, R.~Guarese, and A.~Bergman are with Digital Futures, KTH Royal Institute of Technology, Stockholm, Sweden (e‑mail: michelgk@kth.se; guarese@kth.se; anders.bergman@astrazeneca.com).}%
\thanks{R.~Guarese and M.~Romero are with the School of Electrical Engineering and Computer Science, KTH Royal Institute of Technology, Stockholm, Sweden (e‑mail: \{guarese, marior\}@kth.se).}%
\thanks{F.~Johnson, A.~Bergman, B.~Edvinsson, and J.~Vachier are with AstraZeneca, Stockholm, Sweden (e‑mail: \{fabian.johnson, anders.bergman, benjamin.edvinsson, jeremy.vachier\}@astrazeneca.com).}%
\thanks{X.~V.~Wang is with the School of Industrial Engineering and Management, KTH Royal Institute of Technology, Stockholm, Sweden (e‑mail: wangxi@kth.se).}%
\thanks{M.~Romero is also with the Department of Science and Technology, Linköping University, Norrköping, Sweden (e‑mail: mario.romero@liu.se).}%

}

\begin{document}
\maketitle

\begin{abstract}
We introduce PerfCam, an open source Proof-of-Concept (PoC) digital twinning framework that combines camera and sensory data with 3D Gaussian Splatting and computer vision models for digital twinning, object tracking, and Key Performance Indicators (KPIs) extraction in industrial production lines. By utilizing 3D reconstruction and Convolutional Neural Networks (CNNs), PerfCam offers a semi-automated approach to object tracking and spatial mapping, enabling digital twins that capture real-time KPIs such as availability, performance, Overall Equipment Effectiveness (OEE), and rate of conveyor belts in the production line. 
We validate the effectiveness of PerfCam through a practical deployment within realistic test production lines in the pharmaceutical industry and contribute an openly published dataset to support further research and development in the field. The results demonstrate PerfCam’s ability to deliver actionable insights through its precise digital twin capabilities, underscoring its value as an effective tool for developing usable digital twins in smart manufacturing environments and extracting operational analytics.
\end{abstract}

\begin{IEEEkeywords}
Digital Twins, Machine Learning, 3D Gaussian Splatting, 3D Reconstruction, Object Detection, Computer Vision, Industry 5.0
\end{IEEEkeywords}

\section{Introduction}
\label{sec:intro}

\begin{figure*}[t]
  \centering
  \includegraphics[width=\textwidth]{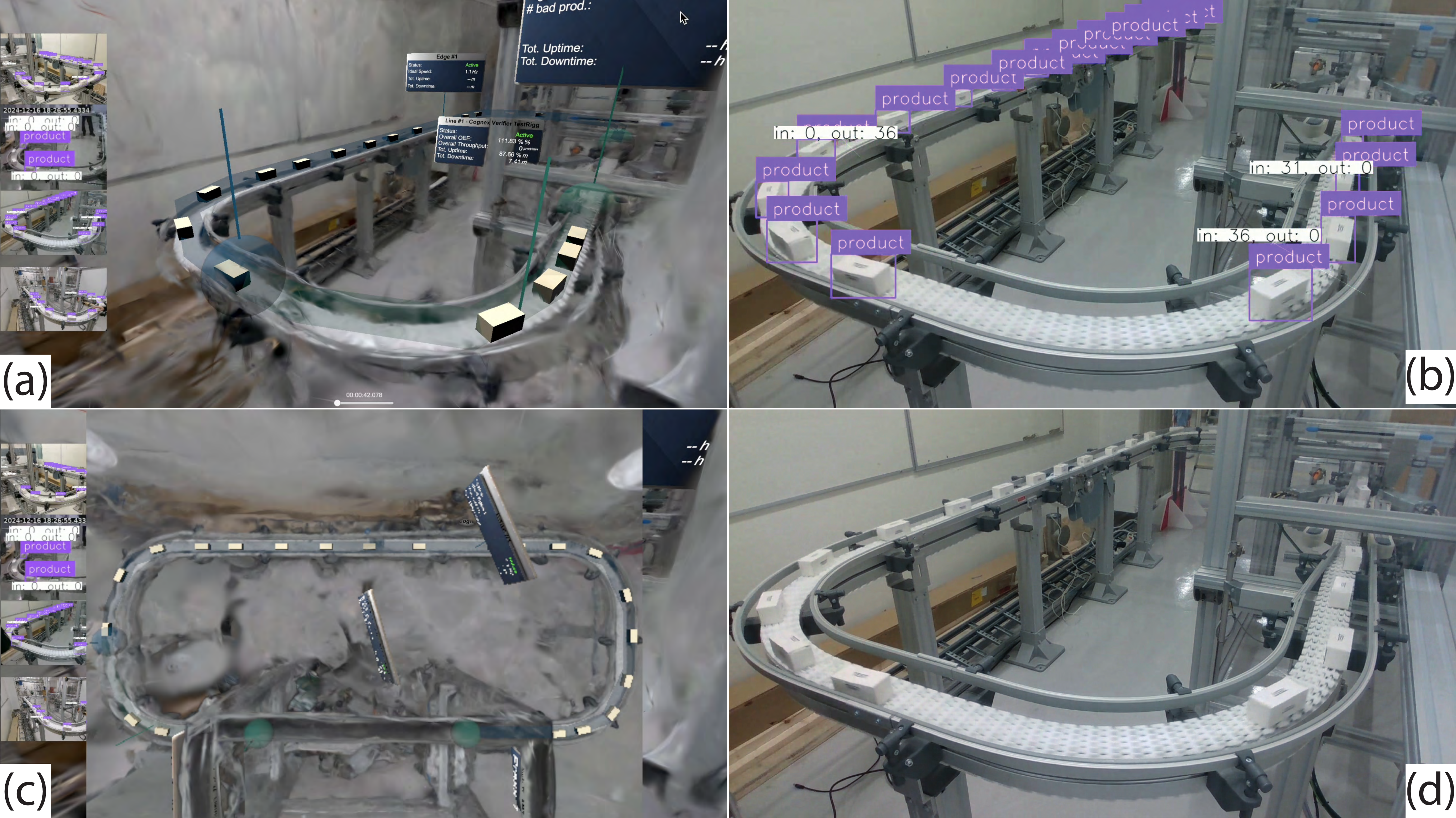}
  \caption{Visual representations in PerfCam vs. ground truth. (a) A 3D reconstruction view based on 3D Gaussian Splatting with a GUI to see various KPIs, 
    (b) a low-latency annotated view of the production line from different camera angles,
    and (c) a 2.5D map of the production line, which reflects the ground truth (d).}
  \label{teaserFigure}
\end{figure*}

    \IEEEPARstart{D}{igital} twins have emerged as an impactful technology for replicating and interacting with complex systems across various domains, including manufacturing~\cite{Weng_2024_CVPR,10.1007/978-3-030-77626-8_35,KRITZINGER20181016,doi:10.1080/0951192X.2022.2027014}, aerospace \cite{9656111,Albuquerque2023}, airport operations \cite{CONDE2022101723}, and urban planning~\cite{LEHTOLA2022102915,Xu2024}, extending into software-intensive sectors such as network infrastructure \cite{HAKIRI2024110350,9854866}, cloud computing~\cite{9429703,9652084}, cybersecurity~\cite{DEAZAMBUJA202425,9566277}, and digital services~\cite{ZHANG201957}. A digital twin is widely known as a virtual representation of a real-world entity, living or nonliving object, process, or system that is continuously updated with real-time data, enabling analysis and optimization of performance \cite{Grieves2017}. The evolution of digital twins has been accelerated by advances in computer vision \cite{9827592}, sensor technology \cite{10.1063/1.5031520,Biegel2024}, and generative artificial intelligence~\cite{WU20243481,10638534}.
    
    In recent years, digital twins had been evolved and advanced in various technologies, disciplines, and industries, leading to development of several levels of details, types, and variations. In this work, we focus on digital twins from a computer vision perspective. Digital twins can be categorized into several hierarchical levels and from various perspectives, each offering varying degrees of detail and functionality \cite{intel2023optimizing}. One of the most fundamental levels are Descriptive Twins \cite{Awonaike_2023,intel2023optimizing}, which provide a basic virtual representation of the physical entity, capturing its static and dynamic states through real-time data synchronization. Building upon this, Informative Twins represent an intermediate level that not only synchronizes real-time sensory and operational data but also integrates contextual and historical information to provide deeper insights into the system's behavior and performance \cite{Wang2024}. Unlike Descriptive Twins, which primarily focus on the current state of the physical entity, Informative Twins incorporate data analytics and visualization tools to analyze trends, identify patterns, and uncover underlying causes of system behaviors. The next level are Predictive Twins \cite{10.1007/978-3-030-61725-7_1,doi:10.2514/6.2020-0418}, which incorporate advanced analytics and machine learning algorithms to forecast future states and behaviors, enabling proactive decision-making and maintenance. The most sophisticated level, Prescriptive Twins, not only predict potential issues but also suggest optimal actions to enhance performance and mitigate risks.
    
    Recent research has focused on enhancing the fidelity and efficiency of digital twin models. High-precision 3D reconstruction techniques are essential for creating accurate digital representations of physical environments~\cite{9757457,10379064}. Traditional methods like Structure-from-Motion (SfM)~\cite{Ullman1979TheIO} and Multi-View Stereo (MVS)~\cite{Wang2024LearningbasedMS} can struggle with reconstructing scenes that have complex geometries or challenging lighting conditions, often resulting in incomplete or less detailed 3D models. To achieve higher-quality and more photorealistic 3D reconstructions, novel approaches such as Neural Radiance Fields (NeRF)~\cite{10.1145/3503250} and 3D Gaussian Splatting~\cite{Kerbl20233DGS} have been proposed, offering improved rendering and representation of scenes.
    
    Simultaneously, the rise of deep learning techniques has led to substantial progress in the fields of object detection and tracking. Algorithms like You Only Look Once (YOLO)~\cite{7780460} and its successors \cite{Redmon2016YOLO9000BF,Redmon2018YOLOv3AI,Bochkovskiy2020YOLOv4OS,Ge2021YOLOXEY,Xu2022PPYOLOEAE,Wang2024YOLOv10RE,Wang2024YOLOv9LW} provide real-time detection capabilities, which are crucial for monitoring moving objects in industrial environments~\cite{make5040083}. Integrating these models into digital twin frameworks enhances the ability to analyze and optimize production processes.
    
    Despite these advancements, challenges remain in deploying digital twin solutions in industrial settings. High costs associated with additional sensor deployment, computational demands, and the complexity of integrating various technologies hinder widespread adoption~\cite{ATTARAN2023100165,9899718}. There is a need for scalable, cost-effective solutions that leverage existing infrastructure while providing accurate and real-time insights.
    
    Although there are multiple methods for building digital twin models—such as incorporating LiDAR-based mapping, CAD-driven simulations, or advanced sensor fusion—camera-based approaches offer significant advantages. Cameras are often more readily available, more cost-effective, and simpler to integrate, especially in settings where minimal additional infrastructure can be introduced. Moreover, camera systems naturally facilitate the integration of object detection and tracking techniques, enabling real-time monitoring of both objects and processes. This synergy accelerates deployment and makes it easier to adapt the system to new production lines or changing environments, ultimately reducing both setup complexity and cost.
    
    In this study, we make four key contributions by adopting a primarily informative approach to digital twinning while also incorporating selected descriptive elements, such as extracting important KPIs. \textbf{First}, we develop \textbf{PerfCam}, an open-source Proof-of-Concept (PoC) digital twinning framework that integrates 3D Gaussian Splatting with real-time object detection to construct precise digital twins of industrial production lines based on visual and sensory data (as shown in Figure~\ref{teaserFigure}).  PerfCam’s current digital twin capabilities include real-time 3D reconstruction, automated object detection and tracking, as well as the ability to extract performance metrics directly from camera feeds. This approach leverages existing camera systems for both 3D reconstruction and object tracking, reducing the need for additional sensors and minimizing initial setup and calibration efforts. \textbf{Second}, we demonstrate that PerfCam enables accurate extraction of Key Performance Indicators (KPIs) such as Overall Equipment Effectiveness (OEE)~\cite{Nakajima1995-gd}, throughput, and conveyor belt speed directly from visual data. This lowers data gathering costs and simplifies scalability across different production lines by eliminating the dependency on specialized hardware, and can also serve as an extra source of sensing for multi-sensor fusion. \textbf{Third}, we incorporate annotated visualization of production line, providing interactive visualizations that aid in identifying bottlenecks and contribute to the optimization of production processes. These visualizations provide operators with a live view of the production line, enriched with annotations and additional KPIs for better monitoring and analysis. \textbf{Fourth}, we validate the system's performance through deployment in a test line within a real-world industrial environment resembling a potential pharmaceutical production line to construct a digital twin, evaluating its effectiveness in practical settings. In addition, we provide an openly published dataset to support further research in this area \cite{73cd-3668-25}.

\section{Related Work}
\label{sec:related_work}

The integration of digital twins, 3D reconstruction, real-time object detection, and annotated visualization has significantly impacted industrial manufacturing and production lines. In this section, we review relevant literature and existing systems that are similar to our proposed platform, particularly focusing on those utilizing cameras mounted around production lines for automating the process of digital twinning.

\textbf{Camera-based Digital Twins in Manufacturing.} Digital twins are virtual representations of physical systems, continuously updated with real-time data to mirror the state of their physical counterparts. In manufacturing, camera-based digital twins facilitate monitoring and optimization of production processes. For instance, Minos-Stensrud \textit{et al.}~\cite{8571914} explored automated 3D reconstruction in Small and Medium Enterprise (SME) factories to facilitate digital twin model generation. They utilized low-cost sensors like RGB-D cameras mounted on drones, employing Simultaneous Localization and Mapping (SLAM) techniques to create 3D models of factory environments. While their approach demonstrated the feasibility of cost-effective digital twin creation, it relied on drones and depth sensors, which may face challenges in cluttered or confined industrial spaces. Chen \textit{et al.}~\cite{CHEN2024100503} proposed a semantic-rich digital twin framework for monitoring and anomaly detection in air-conditioning and mechanical ventilation systems. They employed 3D laser scanning to generate detailed geometric models, integrating them with time-series sensor data to enhance the processes. Although their approach demonstrates the benefits of combining geometric and operational data, it is tailored to static building environments and depends on specialized scanning equipment, which may not be suitable for the dynamic and cluttered conditions of industrial production lines. Zhu and Ji~\cite{Zhu2022} proposed a digital twin-driven method for online quality control in the process industry. Their approach integrates a process simulation model with advanced predictive algorithms to monitor and optimize quality-related parameters in real time. While effective in enhancing predictive accuracy and decision-making capabilities, their method primarily relies on process simulations and data-driven models rather than utilizing camera-based 3D reconstruction.

Alexopoulos \textit{et al.} \cite{doi:10.1080/0951192X.2020.1747642} demonstrated the integration of simulation tools with digital twins to optimize manufacturing workflows, emphasizing the potential for synthetic datasets generated from digital twins to improve machine learning model training in supervised scenarios. Urgo \textit{et al.} \cite{URGO2024249} introduced a method leveraging digital twins to generate labeled synthetic datasets for training Convolutional Neural Networks (CNNs) aimed at object detection in manufacturing environments. This approach addressed the significant challenge of acquiring large and high-quality datasets for training machine learning models by simulating real-world scenarios within a digital twin environment. The proposed methodology allowed the training of CNNs to monitor manufacturing entities such as components, fixtures, and tools by using synthetic images generated within the digital factory twin which significantly reduces the dependency on physical datasets, enabling efficient model development even in inaccessible or non-existent systems. However, their approach heavily relies on pre-configured digital twins, requiring substantial manual effort to set up and maintain the synthetic data generation pipeline. Moreover, the lack of seamless automation in integrating real-time data into the digital twin limits its adaptability to rapidly changing industrial environments. This manual configuration and dependence on static simulation scenarios reduce the flexibility and scalability of the solution in dynamic manufacturing systems.

Dynamic camera position optimization, as presented by Wang \textit{et al.}, leverages digital twins to ensure comprehensive coverage of production environments using 3D Gaussian splatting and reinforcement learning, thereby improving object visibility and detection rates~\cite{doi:10.1080/00207543.2023.2252108}. Moreover, the integration of real-time vision-based systems into digital twins enables automated safety checks and enhances operator interaction through Augmented Reality (AR) interfaces, providing actionable insights directly to operators in the production line~\cite{UHLEMANN2017335,10275696}.

These advancements highlight how camera-based digital twins, enriched with AI and real-time data integration, offer scalable, cost-effective solutions for modern manufacturing environments, ultimately driving innovation in monitoring, process optimization, and resource management.

\textbf{Integration of 3D Reconstruction with Object Detection.} Combining 3D reconstruction with camera-based object and event detection methods enhances the accuracy and reliability of monitoring systems. FusionVision presents a comprehensive approach to 3D object reconstruction and segmentation from RGB-D cameras, integrating state-of-the-art object detection techniques with advanced instance segmentation methods~\cite{s24092889}. This integration enables a holistic interpretation of RGB-D data, facilitating the extraction of comprehensive and accurate object information. Similarly, AutoRecon introduces a framework for automated 3D object discovery and reconstruction, utilizing multi-view images and integrating object localization and segmentation techniques~\cite{wang2023autorecon}. These approaches demonstrate the potential of integrating 3D reconstruction with object detection to improve monitoring and inspection processes in manufacturing.

\textbf{KPI Extraction Using Industrial Vision Systems.} Vision systems enable the extraction of KPIs by analyzing visual data to monitor production metrics such as throughput, speed, and equipment effectiveness. For example, IVC's high-speed video monitoring system \cite{ivc_high_speed_video_system} captures video of high-speed processes, allowing frame-by-frame playback to help identify and resolve issues quickly, thereby optimizing production line performance. Similarly, advanced AI-driven vision platforms such as NeuroSpot™ ~\cite{neurospot} provide real-time analytics capabilities by integrating AI-powered video analytics into industrial workflows. These systems analyze surveillance footage to deliver actionable insights, enabling operators to enhance operational efficiency and improve decision-making. By monitoring and extracting KPIs such as customer traffic patterns, table occupancy and turnover rates, staff performance metrics, and patient safety compliance, NeuroSpot™ demonstrates the versatility of vision-based analytics in various operational contexts. These capabilities highlight the potential of such vision-based monitoring systems to transform data-rich environments into optimized, efficient workflows tailored to specific needs.

\setcounter{figure}{1} 
\begin{figure*}[t] \centering \includegraphics[width=1\linewidth]{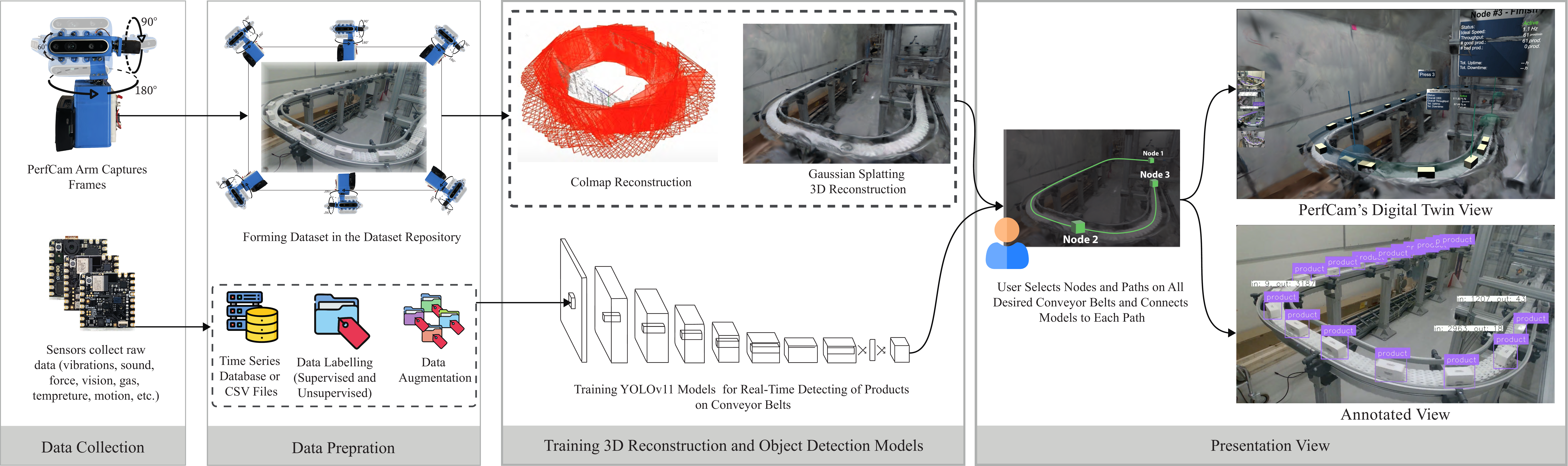} \caption{System workflow of PerfCam. PerfCam uses 4-6 small robotic arms with RGB cameras to capture frames around the production line, while sensors record multimodal data in a time-series DB. It reconstructs a detailed 3D model using COLMAP and Gaussian Splatting and then employs a CNN-based object detection and tracking to extract KPIs such as conveyor belt throughput, number of stops, and OEE. Finally, PerfCam provides 3D reconstructed and annotated views, with annotated visualizations enabling users to monitor and improve production KPIs and processes interactively.} \label{fig:architecture} \end{figure*}

\section{Method} \label{sec}

In this section, we present the architecture and components of the PerfCam framework, which integrates COLMAP 3D reconstruction for SfM \cite{schoenberger2016sfm} and MVS \cite{schoenberger2016mvs} generation and then 3D Gaussian Splatting based on the Surface-Aligned Gaussian Splatting (SuGaR) method \cite{10655755} for high-fidelity scene reconstruction with real-time object detection and tracking using YOLO-based models \cite{Wang2024YOLOv10RE,yolo11_ultralytics}. The system is designed to extract KPIs such as conveyor belt throughput, number of stops, and OEE in industrial production lines directly from visual data to provide real-time actionable information for optimizing production processes.

\subsection{System Workflow} 

Figure~\ref{fig:architecture} shows the complete workflow of PerfCam. The process begins with \emph{Image Acquisition}, where images are captured using RGB cameras mounted on robotic arms. Next, in the \emph{3D Reconstruction} phase, a point cloud is generated using COLMAP and enhanced with 3D Gaussian Splatting. Following this, during \emph{Object Annotation}, frames and products are labeled using supervised or unsupervised methods to create an annotation set. In the \emph{Model Training} step, the YOLOv11-based model is fine-tuned. Finally, in \emph{Real-Time Detection}, the trained model is deployed to detect and track objects in real time.

\textbf{Data Collection and Preprocessing.} The data collection process involves capturing a series of images from multiple viewpoints around the production line using PerfCam's camera module.

We developed a camera module for PerfCam using an Intel RealSense D435i depth camera\cite{IntelRealSenseD400}, mounted on a compact three-axis robotic arm (as shown in the Data Collection step in Figure~\ref{fig:architecture}). We used this module for both collecting data for the 3D reconstruction phase as well as for the object detection phase. This module is driven by a Raspberry Pi 5\cite{RaspberryPi5Brief}, which controls three MG995 servo motors \cite{MG995Datasheet}. In this PoC, the camera captures frames using the Raspberry Pi during the 3D reconstruction phase. For the object detection phase, it is directly connected to a computer to enhance performance. The base, middle, and top servo motors enable rotations of $180^\circ$, $90^\circ$, and $60^\circ$ in the yaw, roll, and pitch axes respectively.

In addition to visual data, PerfCam collects sensory data from various sensors deployed around the production line. These sensors measure parameters such as vibrations, sound, force, gas levels, temperature, and motion.

The sensory data can be either stored in a time-series database such as InfluxDB, facilitating efficient storage, retrieval, and real-time analysis or in a simple Comma-Separated Value (CSV) format. This data is integrated with the visual data based on synchronized timestamps, enabling comprehensive monitoring of the production line.

\textbf{3D Reconstruction with COLMAP and 3D Gaussian Splatting.} The captured images are first processed using COLMAP~\cite{schoenberger2016sfm,schoenberger2016mvs} to perform SfM and MVS, generating a sparse and then dense point cloud representation of the scene.

\textbf{Object Annotation and Data Labelling for Model Training.} We proceed with data labelling to create ground truth annotations necessary for training the object detection model. Each augmented image is annotated to identify the objects of interest, resulting in a set of labelled bounding boxes. The labelling process involves the following steps:

\begin{enumerate}
    
\item \emph{Manual Annotation:} Human annotators utilize specialized labeling software \cite{labelstudio} to manually outline objects with bounding boxes around the images and assign class labels. This ensures high-quality annotations.

\item \emph{Assisted Annotation:} To expedite the labelling process, we employ semi-automated techniques such as out of the box label assists, like Roboflow Universe label assist \cite{roboflow_universe_label_assist}, as well as segmentation algorithms, like SAM 2 \cite{ravi2024sam2} to delineate objects from the background.

\end{enumerate}

\textbf{Data Augmentation.} After data labelling and to enhance the robustness of the object detection model, we perform data augmentation on the collected images. This results in an augmented dataset increasing the diversity of training samples and improving the model's generalization.

\textbf{Training the YOLOv11-based Model.} With the annotated and augmented dataset, we fine-tune a YOLOv11-based CNN model~\cite{Wang2024YOLOv10RE} for object detection.

\textbf{Real-Time Object Detection and Tracking.} The trained model is deployed for real-time object detection on incoming video streams. Detected objects are tracked across frames using data association techniques. Tracking involves assigning consistent identifiers to objects over time, enabling the computation of trajectories. We integrate tracking and annotation visualizations using the ByteTrack method~\cite{10.1007/978-3-031-20047-2_1} for marker-based tracking.

\subsection{KPI Extraction} 

KPIs are extracted by analyzing the tracked object trajectories and analyzing sensory data. 
The performance metrics are clearly defined in quantitative terms to enable accurate monitoring.

\textbf{Throughput.} represents the number of objects passing a reference point per unit time.

\textbf{Conveyor Belt Speed.} determines how far an object travels within a specified time frame.

\textbf{Downtime Calculation.} Downtime, denoted as $T_{\text{downtime}}$, is the total time during which the production line is not operational due to machine stops or failures. It is calculated based on sensory data indicating machine status. Let $\{ t_{\text{stop}}^n, t_{\text{start}}^n \}_{n=1}^{N_{\text{stops}}}$ denote the start and end times of each downtime event, where $N_{\text{stops}}$ is the total number of stops detected. Then, the total downtime is:

\begin{equation}
T_{\text{downtime}} = \sum_{n=1}^{N_{\text{stops}}} (t_{\text{start}}^n - t_{\text{stop}}^n).
\end{equation}

\textbf{Counting Good and Total Pieces.} The number of good pieces, denoted as \( Q_{\text{good}} \), is determined using any data source or API, e.g., dataset from a laser counting sensor, which tracks products meeting quality standards as they pass a specific point on the production line, or through object detection and tracking data. We define the total number of good pieces as \(Q_{\text{good}}\) and the total
number of defective pieces as \(Q_{\text{bad}}\), respectively. The total number of
pieces is denoted by \(Q_{\text{total}}\). In some cases, \( Q_{\text{bad}} \), is available, allowing the total or good pieces to be calculated using the following relationship:

\begin{equation}
Q_{\text{total}} = Q_{\text{good}} + Q_{\text{bad}}
\end{equation}

\textbf{Ideal Cycle Time and Planned Production Time.} The \emph{Ideal Cycle Time}, denoted as $\tau_{\text{ideal}}$, is the theoretical minimum time required to produce one piece—meaning the process time under perfect conditions with no downtime or inefficiencies—and is provided by the operator based on production specifications. The \emph{Planned Production Time}, denoted as $T_{\text{planned}}$, is the total scheduled production time and is also provided by the operator.

\textbf{Operating Time.} The \emph{Operating Time}, denoted as $T_{\text{operating}}$, is the actual production time when the equipment is operational:

\begin{equation}
T_{\text{operating}} = T_{\text{planned}} - T_{\text{downtime}}.
\end{equation}

\textbf{Overall Equipment Effectiveness.} One of the key metrics is the OEE \cite{Nakajima1995-gd}, which is a comprehensive metric used to evaluate the degree to which a manufacturing operation is used efficiently. OEE combines three critical factors—availability, performance, and quality—into a single measure, providing insights into the efficiency and productivity of the equipment. It is essential for identifying areas of improvement and optimizing production processes.

\begin{equation}
\text{OEE} = \text{Availability} \times \text{Performance} \times \text{Quality},
\end{equation}

Each component is calculated as follows:

- \emph{Availability} ($A$) is calculated as:

\begin{equation}
A = \frac{T_{\text{operating}}}{T_{\text{planned}}} = \frac{T_{\text{planned}} - T_{\text{downtime}}}{T_{\text{planned}}}
\end{equation}

- \emph{Performance} ($P$) is calculated as:

\begin{equation}
P = \frac{\tau_{\text{ideal}} \times Q_{\text{total}}}{T_{\text{operating}}}
\end{equation}

- \emph{Quality} ($Q$) rate is:

\begin{equation}
Q = \frac{Q_{\text{good}}}{Q_{\text{total}}}
\end{equation}

Therefore, the OEE can be expressed as:

\begin{multline}
\text{OEE}=A \times P \times Q=\\
\left( \frac{T_{\text{operating}}}{T_{\text{planned}}} \right) \times \left( \frac{\tau_{\text{ideal}} \times Q_{\text{total}}}{T_{\text{operating}}} \right) \times \left( \frac{Q_{\text{good}}}{Q_{\text{total}}} \right)
\end{multline}

 By continuously monitoring these KPIs, PerfCam enables spotting the production line constraints and wasteful practices in the digital twin. The real-time data allows for immediate corrective actions, optimizing overall operational efficiency.

 PerfCam is designed for ease of deployment in real-world industrial settings. The use of existing camera infrastructure minimizes additional costs, and the platform's scalability ensures it can adapt to different production line configurations.

 To support further research, we provide an openly published dataset comprising the captured images, annotations, and reconstructed 3D models. This dataset is a useful resource for improving digital twinning and KPI extraction methodologies in future researches.

\subsection{PerfCam's Digital Twin View and Deployment}

\textbf{Tracking and Annotated Visualization.}  By placing ByteTrack's bounding boxes in the environment, we establish reference coordinates that allow for the overlay of virtual KPIs and object trajectories onto live video feeds, in a situated visualization fashion \cite{leeSituatedVis2023}, which has been shown to positively affect decision-making \cite{zheng2024DecisionMaking}. This provides operators with an interactive and intuitive means to monitor production metrics, as shown in Figure \ref{teaserFigure}-B.

\textbf{Visualization of Sensory Data in GUI.} The GUI integrates the sensory data from the time-series database or files, providing real-time charts and dashboards that display parameters such as temperature, vibration levels, and other environmental metrics. Operators can view historical trends and correlate sensory data with production metrics, facilitating informed decision-making \cite{electronics10070828}.

\textbf{Deployment in Industrial Environment.} PerfCam, even though currently a research PoC, is purposefully designed for seamless deployment in real-world industrial environments. By leveraging potentially existing camera infrastructure in the production environment, it minimizes additional costs, while its modular design allows for easy adaptation to various production line configurations and integration with existing industrial systems.

\textbf{Dataset and Code Contributions.} To support further research, we provide an openly published dataset comprising the camera footages, annotations, sensory data, and reconstructed 3D models. Additionally, we have bundled both the PoC Raspberry Pi and Unity projects used to conduct the experiments and develop the digital twin presented in this work.

\begin{figure*}[t]
    \centering
    \includegraphics[width=\linewidth]{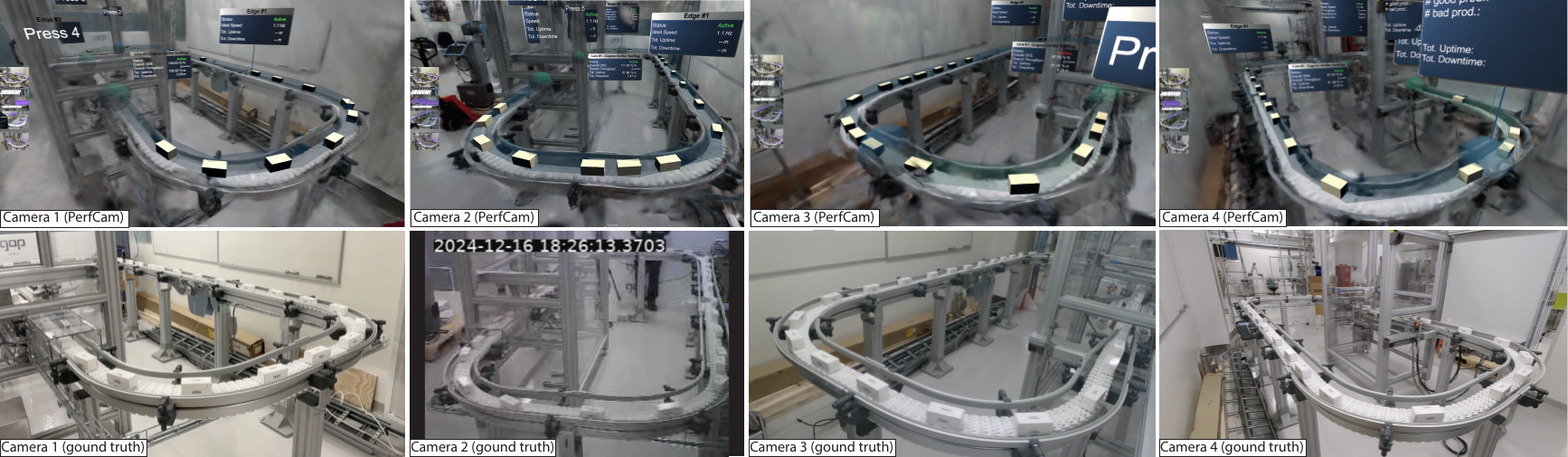} %
    \caption{Snapshots from video footage captured by Cameras 1–4 during the experiments on the test line (second row) alongside corresponding views in the PerfCam's digital twin (first row).}
    \label{fig:perfcam-view}
\end{figure*}

\begin{figure*}[b]
\centering
\begin{subfigure}{0.32\linewidth}
  \centering
  \includegraphics[width=\linewidth]{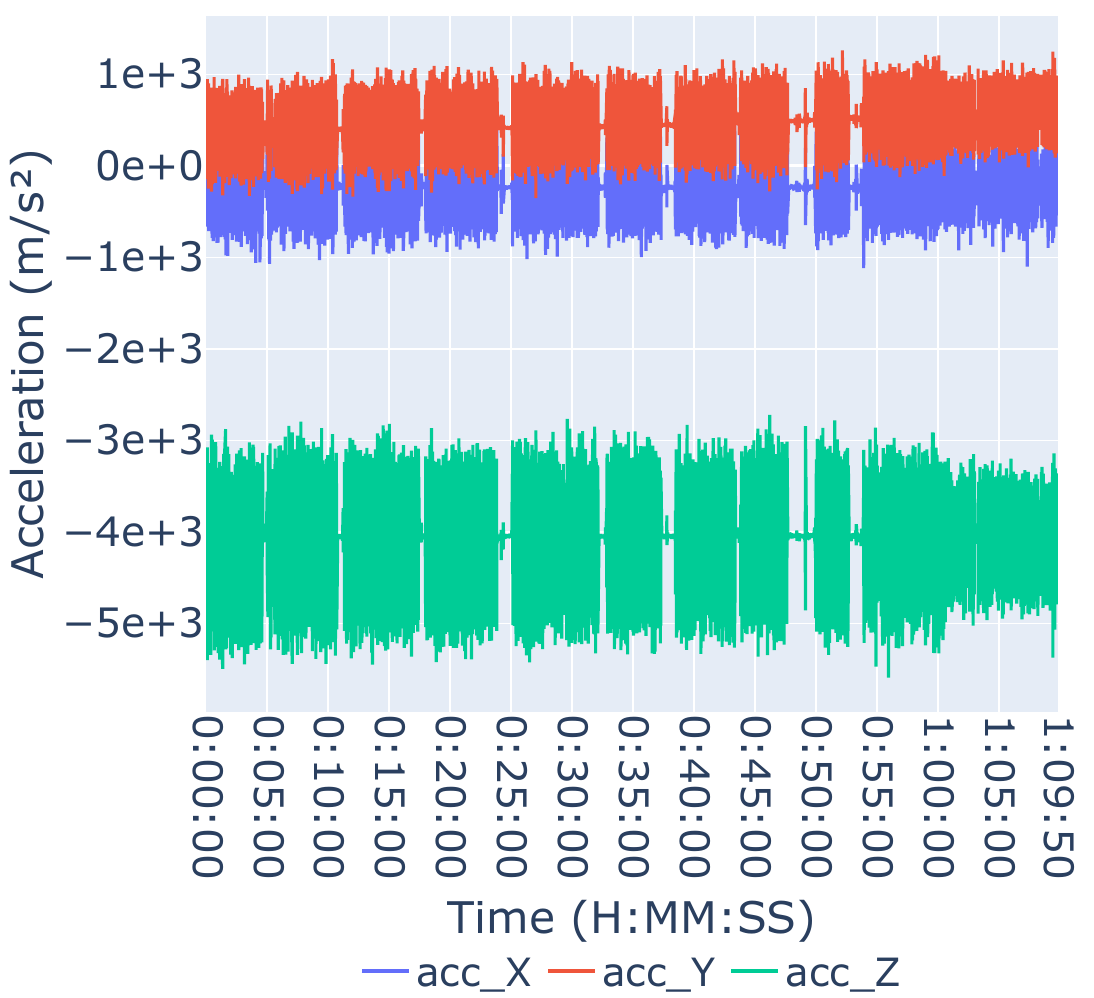}
  \caption{Acceleration readings over time}
  \label{fig:accel}
\end{subfigure}
\hspace{0.001\linewidth}
\begin{subfigure}{0.32\linewidth}
  \centering
  \includegraphics[width=\linewidth]{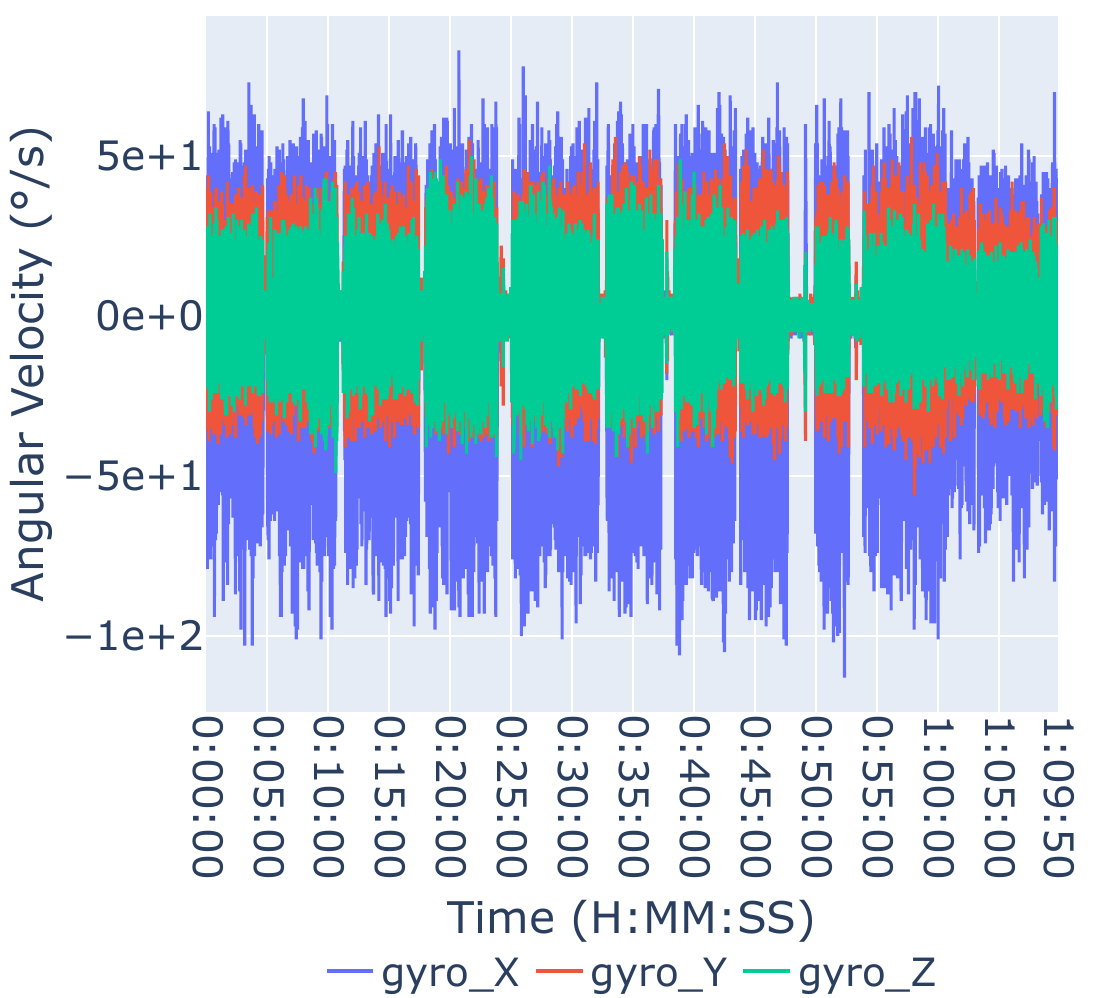}
  \caption{Gyroscope readings over time}
  \label{fig:gyro}
\end{subfigure}
\hspace{0.001\linewidth}
\begin{subfigure}{0.32\linewidth}
  \centering
  \includegraphics[width=\linewidth]{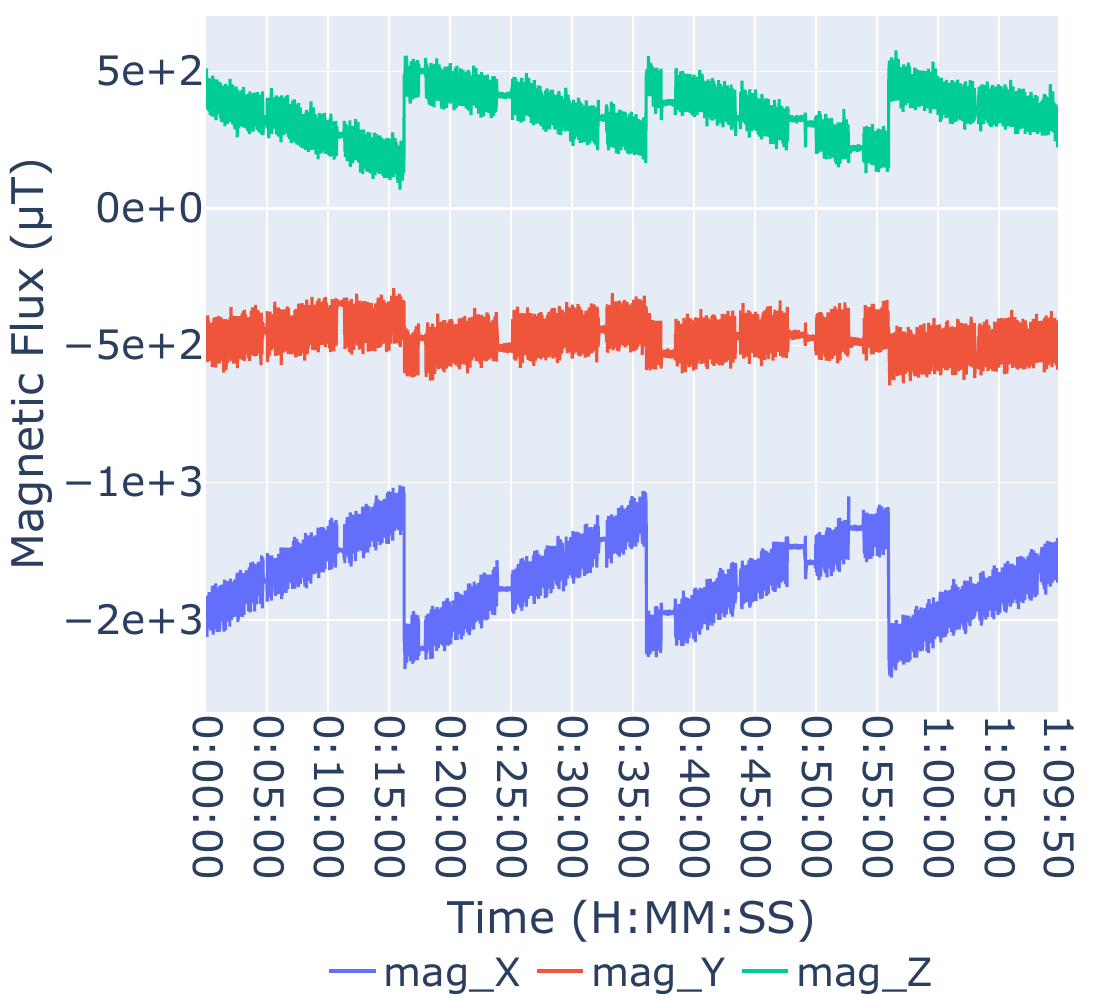}
  \caption{Magnetometer readings over time}
  \label{fig:mag}
\end{subfigure}

\caption{Sensor measurements from the motor.}
\label{fig:sensor_plots}
\end{figure*}
\setcounter{figure}{5}
\begin{figure*}[b]
    \centering
    \includegraphics[width=\linewidth]{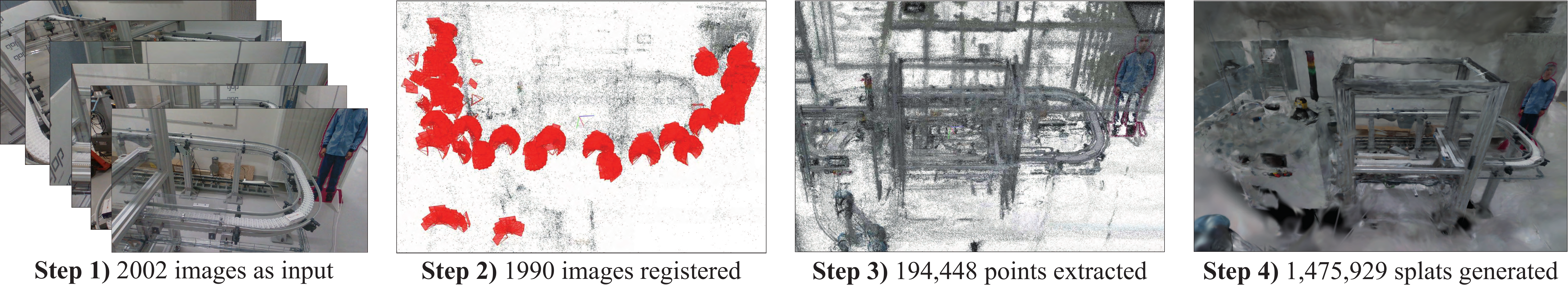}
    \caption{Step-by-step process of 3D reconstruction. Starting from raw camera footage (step 1), progressing to successfully registered images in COLMAP (step 2), transitioning to the extracted point cloud (step 3), and resulting in the generation of Gaussian splats (step 4).}
    \label{fig:perfcam-reconstruction}
\end{figure*}
\setcounter{figure}{4}
\begin{figure}[t]
  \centering
  \includegraphics[width=\linewidth]{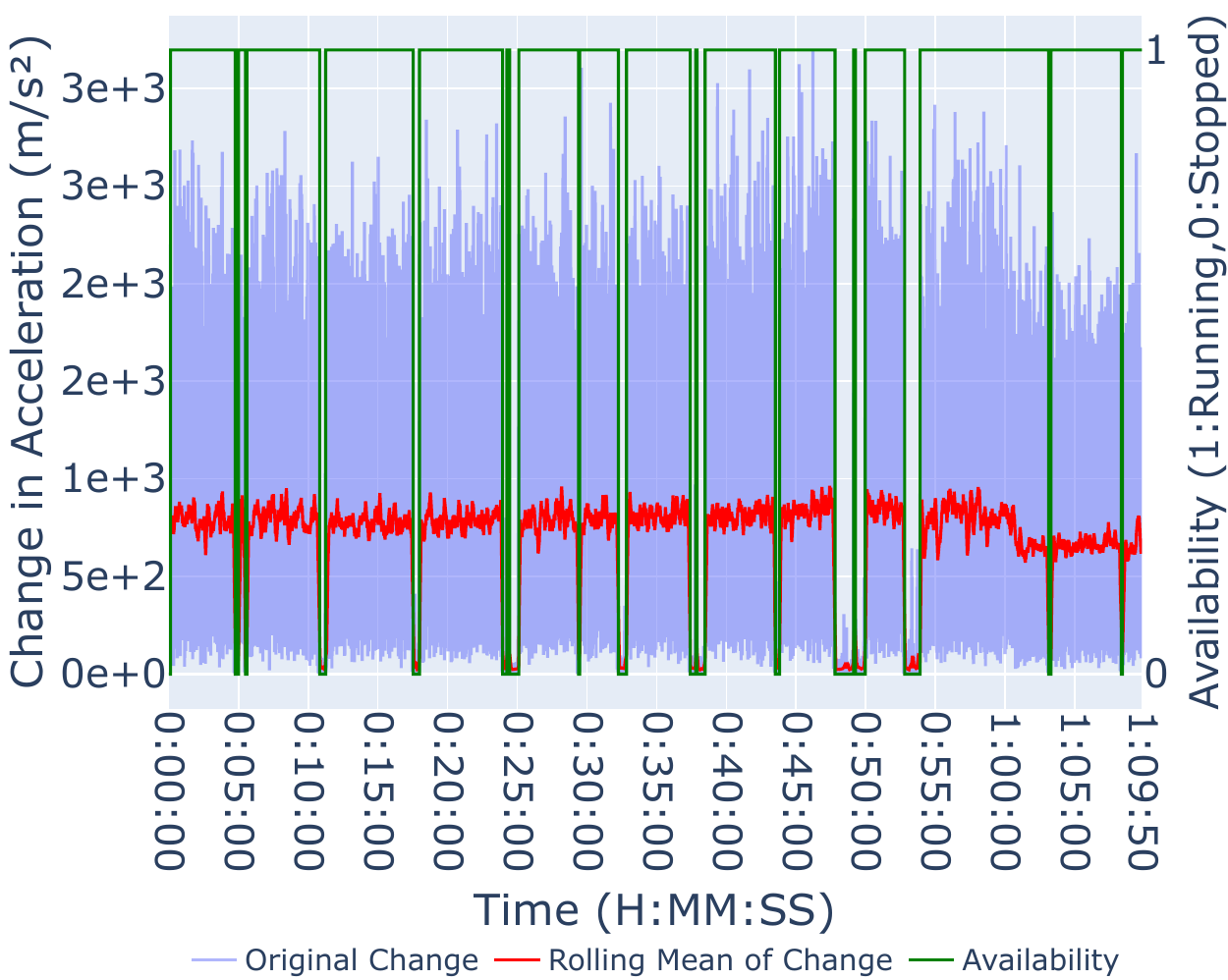}
  \caption{Availability (in \%) of conveyor belt calculated based on the changes in motor's acceleration.}
  \label{fig:avail_calc}
\end{figure}
\setcounter{figure}{6}

\section{Experiments}
\label{sec:experiments}
In this section, we describe the single experiment conducted to evaluate PerfCam in a test production line scenario. Our goal is to demonstrate the system's ability to (1) reconstruct a 3D model of the line, (2) detect and track moving products, and (3) estimate manufacturing metrics such as OEE, availability, and product counts. We first outline the setup and data collection process, and then summarize the main steps undertaken during the experiment.

\subsection{Experiment's Scenario}

We deployed a simplified circular conveyor belt in a test environment that mimics an industrial production line. The conveyor belt is configured in an infinite loop with three designated nodes, shown in Figure \ref{fig:architecture}:

\begin{itemize}
    \item \textbf{Node 1 (Start):} Products or raw material enter the conveyor line here.
    \item \textbf{Node 2 (Quality Assurance):} A human operator inspects products in real time. If a product is faulty, the operator removes it from the line. If needed, the operator also repositions any slightly misaligned products.
    \item \textbf{Node 3 (End):} Final products exit the line for packaging or shipping.
\end{itemize}

Though physically realized as a single closed-loop conveyor, these nodes represent the core steps in a standard production process—\emph{input of raw materials}, \emph{inspection/adjustment}, and \emph{final output}. By design, each node helps validate whether PerfCam can:

\begin{enumerate}
    \item Recognize and track products moving on the conveyor.
    \item Detect occasional machine stops (i.e., downtime).
    \item Monitor operator-driven interventions (removal of faulty items).
    \item Calculate and visualize KPIs such as availability, performance, quality, and OEE.
\end{enumerate}

Over the course of the experiment, products were periodically placed on the line at \textbf{Node 1}. At \textbf{Node 2}, an operator performed a basic quality check. If a product was deemed faulty, it was removed entirely; otherwise, it continued along to \textbf{Node 3}, simulating an end-of-line packaging step. The entire process was repeated, creating a continuous flow of items through the loop. Periodic stoppages were also introduced to mimic common real-world downtime scenarios, and PerfCam tracked these to determine \textit{availability}. We further used \textit{performance} and \textit{quality} measurements, along with the \emph{ideal cycle time} set to 1.1 seconds ($\tau_{\text{ideal}}=1.1\,\mathrm{s}$), to compute the \emph{OEE}. This 1.1-second value was derived from analysing the video footage of the process under optimal operating conditions, where we measured the interval between two consecutive boxes passing node 3 while maintaining a buffer of at least two empty boxes between them.

\subsection{Experimental Setup}

\subsubsection{Camera Configuration}
To capture the conveyor from multiple angles and test various resolutions, four cameras were arranged around the conveyor belt (as shown in Figure \ref{fig:perfcam-view}). Table~\ref{tab:camera_spec} summarizes the main characteristics of these devices and corresponding Frames per Second (FPS).

\begin{table}[htbp]
\caption{Camera Specifications in the Test Setup}
\label{tab:camera_spec}
\centering
\begin{tabular}{l|l|l|l|l}
\hline
\textbf{\#} & \textbf{Resolution} & \textbf{Rate} & \textbf{Length} & \textbf{Model} \\
\hline
1 & 1620$\times$1080 & $\sim$29.7\,FPS & 4206s & FaceTime HD \\
2 & 320$\times$240 & $\sim$0.47\,FPS & 3119s & Arduino Nicla Vision \\
3 & 1280$\times$720 & $\sim$20.2\,FPS & 4177s & Intel RealSense D435i  \\
         & 848$\times$480 & $\sim$22.8\,FPS & 4177s & Depth Footage of  D435i \\
4 & 1920$\times$1080 & $\sim$30.1\,FPS & 4204s & Google Pixel\,7 Pro \\
\hline
\end{tabular}
\end{table}

Each camera feed was processed by our YOLO-v11 based object detection module, which generated bounding boxes around products on the belt. In this experiment, we processed the camera footage as an independent step and did not perform the analysis in real-time. Although all four cameras were started and synchronized simultaneously, their recorded lengths differ (as shown in Table \ref{tab:camera_spec}). For instance, Camera~2 stops near the end of minute 52, Camera~3 stops midway through minute 70, while Cameras~1 and~4 stop in minute 71.  This variation reflects real-world scenarios where certain cameras may fail or stop prematurely. Having overlapping coverage ensures that, if one camera feed ends or fails, we still retain data from other cameras. By combining views from these four angles, we mitigated occlusions and tested how PerfCam handles varying resolutions and frame rates.

\subsubsection{Sensor Integration}
In parallel, we attached an Arduino Nicla Sense ME sensor \cite{NiclaSenseME} directly on top of the conveyor’s motor. It continuously recorded:
\begin{itemize}
    \item Acceleration ($m/s^2$) - on 3-Axis X,Y,Z
    \item Angular velocity ($^\circ/s$) - on 3-Axis X,Y,Z
    \item Magnetic flux ($\mu$T) - on 3-Axis X,Y,Z
    \item Ambient pressure (hPa)
    \item Temperature (°C), 
    \item Relative humidity (\%RH), and
    \item CO$_2$ concentration (ppm).
\end{itemize}

These measurements were timestamped in sync with the video streams, enabling PerfCam to correlate sensor fluctuations (e.g., vibrations) with production-line availability (i.e., detecting when the line is running or stopped). This integration of sensory data adds value in two key ways: first, it aids in identifying nuanced correlations that might influence production outcomes, such as how temperature fluctuations or increased vibrations align with machine stoppages. Second, it provokes user engagement by offering insights into how environmental and operational metrics interact. For example, in a digital twin context, these metrics can be visualized and explored when users interact with specific machinery, such as the motor, fostering a deeper understanding of operational dynamics. Figure~\ref{fig:sensor_plots} presents a range of sensor measurements from the motor—namely, acceleration from accelerometer data in Figure~\ref{fig:accel}, angular velocity from gyroscope readings in Figure~\ref{fig:gyro}, magnetic flux from magnetometer readings Figure~\ref{fig:mag}—alongside the derived availability metric in the second Y-axis (right side) of Figure~\ref{fig:avail_calc}, which is calculated from the accelerometer data. This metric assumes a binary form: a value of 1 indicates the line is running, whereas 0 signifies that it is stopped. Together, these diverse sensory inputs emphasizing how diverse sensory inputs contribute to a holistic picture of production health.

\begin{figure*}[t]
\centering
\includegraphics[width=0.95\linewidth]{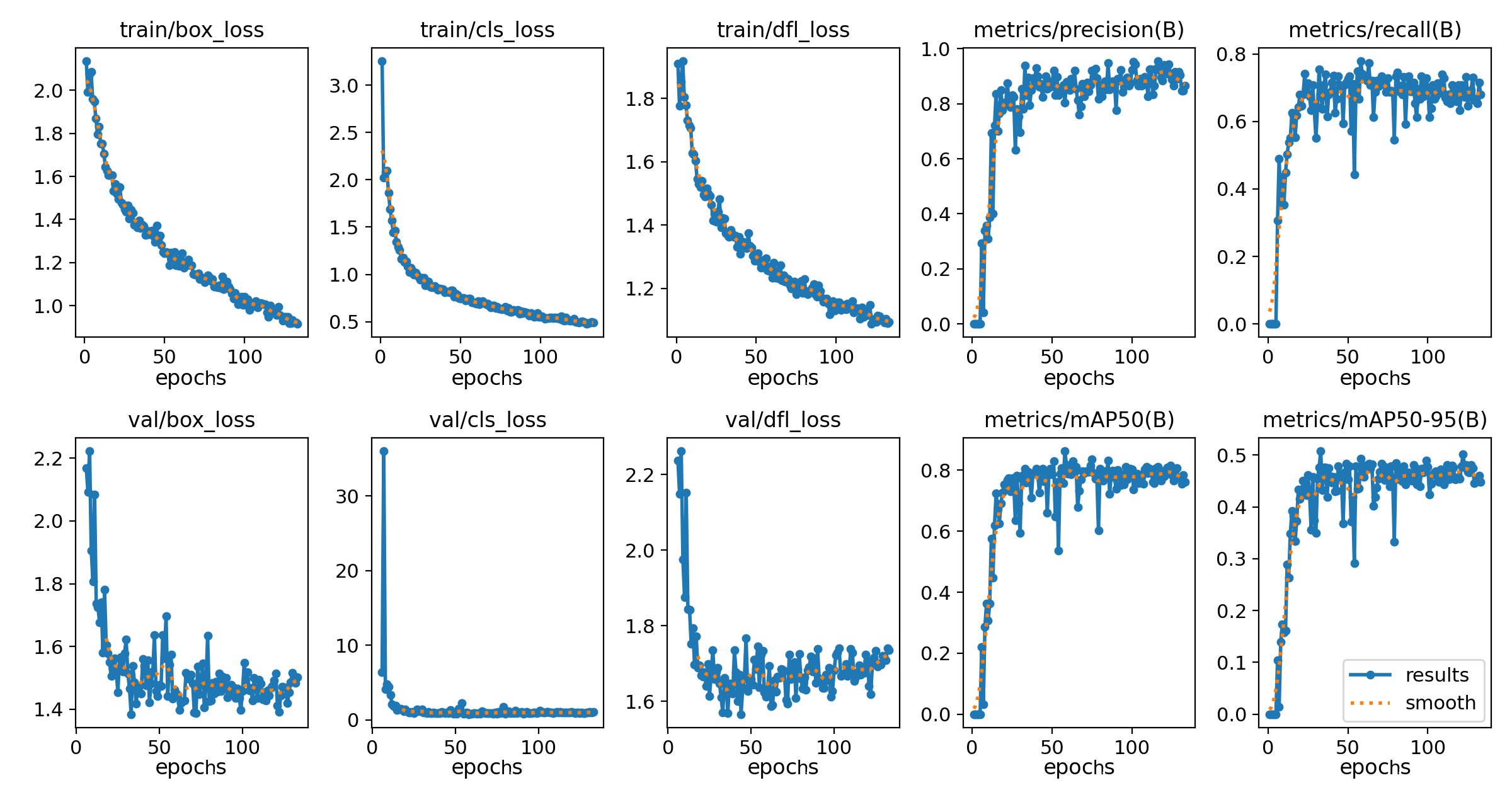}
\caption{
Training and validation metrics for the model.
(a) bounding box regression (lower=better localization).
(b) classification loss (declining=improved classification).
(c) distribution focal loss for finer localization.
(d) \emph{precision(B)}: fraction of predicted positives that are correct (higher=fewer false positives).
(e) \emph{recall(B)}: fraction of actual positives detected (higher=fewer misses).
(f) \emph{mAP50(B)}: mean Average Precision at IoU=0.50.
(g) \emph{mAP50-95(B)}: mean Average Precision from IoU=0.50 to 0.95.
}
\label{fig:results_training}
\end{figure*}

\subsubsection{3D Reconstruction Procedure}

Before continuously monitoring the products, we first created a 3D representation (digital twin) of the conveyor loop. We placed PerfCam's camera module in 26 unique positions around the test line, and at each position, we captured $77$ high-resolution images, resulting in a total of $2002$ RGB frames. Depth information and inertial data (motor angles, IMU readings) were simultaneously logged, allowing us to later cross-check camera poses. Figure \ref{fig:perfcam-reconstruction} represents a step-by-step process:

\begin{enumerate}
    \item \textbf{Image Collection:} First, we positioned PerfCam's camera module at 26 distinct locations surrounding the conveyor, capturing 77 images at each spot. In total, 2002 high-resolution RGB images (plus depth frames) were recorded. Each capture also included servo motor angles and IMU data.
    
    \item \textbf{SfM and MVS Processing:} Next, we utilized COLMAP’s SfM and MVS pipelines to generate registered 1990 images and extract an initial 3D point cloud containing 194,448 points.
    
    \item \textbf{Gaussian Splatting:} Then, we refined the initial point cloud using the SuGaR-based 3D Gaussian Splatting approach to enhance geometric and photometric fidelity.
    
    \item \textbf{Digital Twin Initialization:} Finally, we established the reconstruction as the baseline “digital twin” of the conveyor system. After completion, we overlaid real-time product detection and tracking information onto this virtual environment using the Unity Engine as the base platform for developing and managing the 3D digital twin.
\end{enumerate}

\subsection{Object Detection and Tracking}

Once the 3D environment was established, we monitored the movement of products (white boxes) on the circular belt. Using the four-camera setup described above, we ran a YOLOv11-based model to detect and track products over time. Our detection pipeline generated bounding boxes for each product in each frame, providing temporal trajectories from which we derived throughput and accuracy metrics.

\begin{figure}[htbp]
\centering
  \includegraphics[width=.95\linewidth]{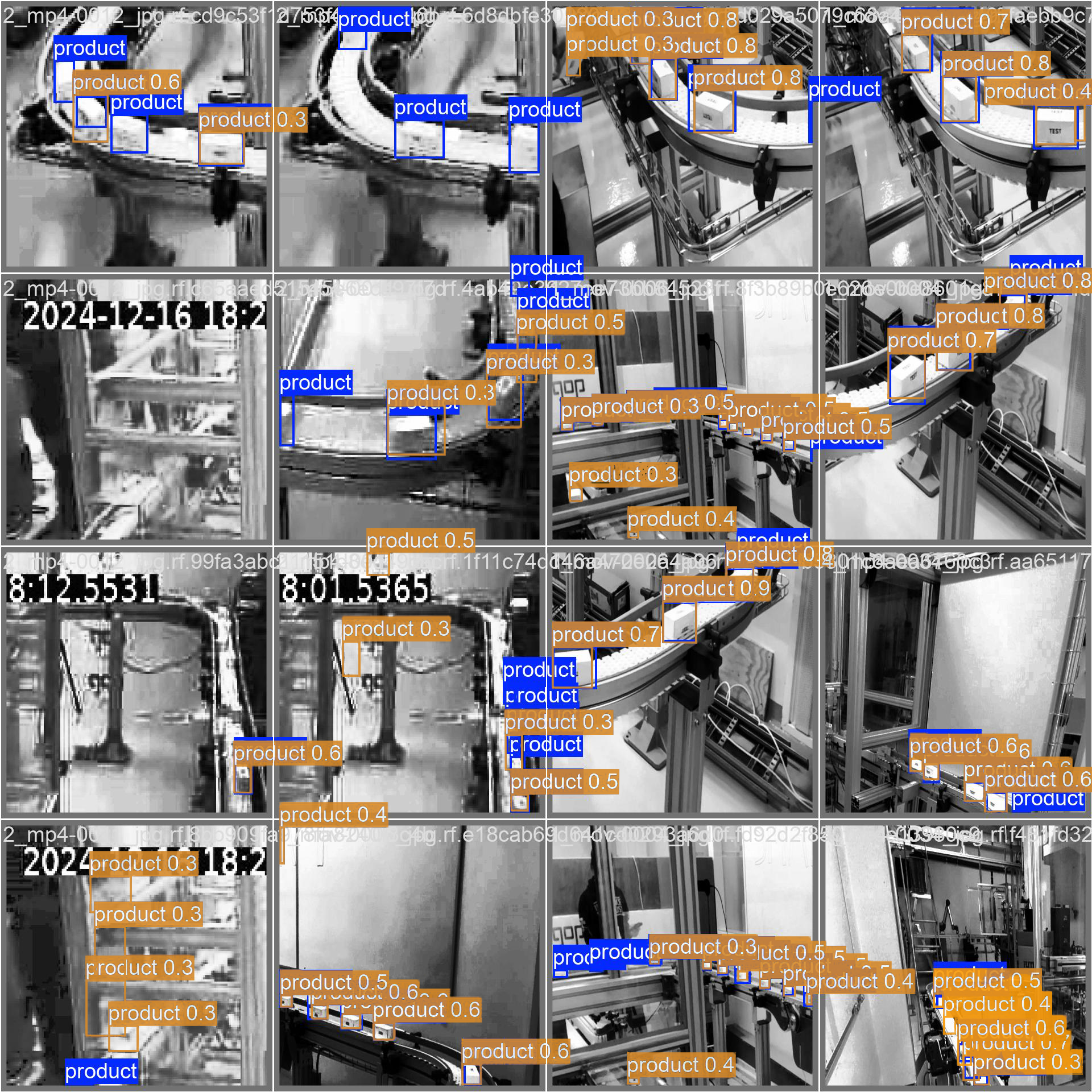}
  \label{fig:val_labels}
\caption{A validation batch. (blue)~Ground truth labels, and (orange)~corresponding predictions by PerfCam.}
\label{fig:validation_images}
\end{figure}

\begin{figure*}[b]
\centering
\begin{subfigure}{0.42\linewidth}
  \centering
  \raisebox{0pt}[\dimexpr\height][\depth]{
    \includegraphics[width=\linewidth]{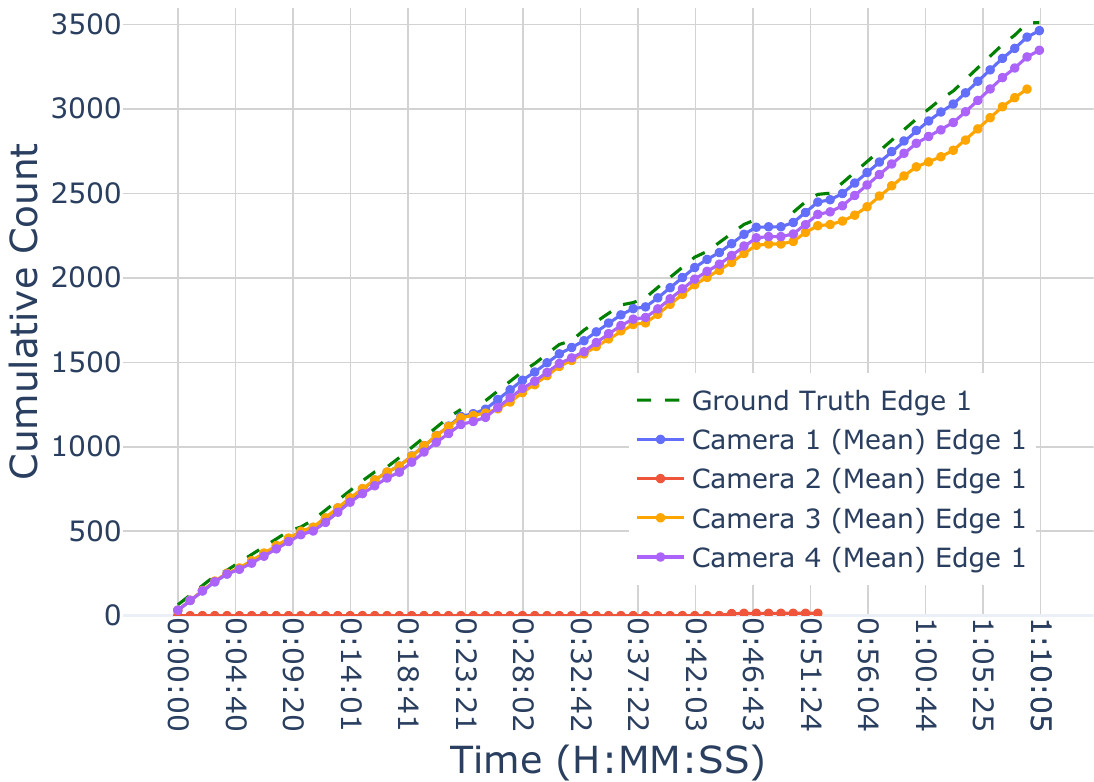}
  }
  \caption{Cumulative Count (Edge 1)}
  \label{fig:cumul1}
\end{subfigure}
\hspace{0.001\linewidth}
\begin{subfigure}{0.43\linewidth}
  \centering
  \raisebox{-2.2pt}[\dimexpr\height][\depth]{
    \includegraphics[width=\linewidth]{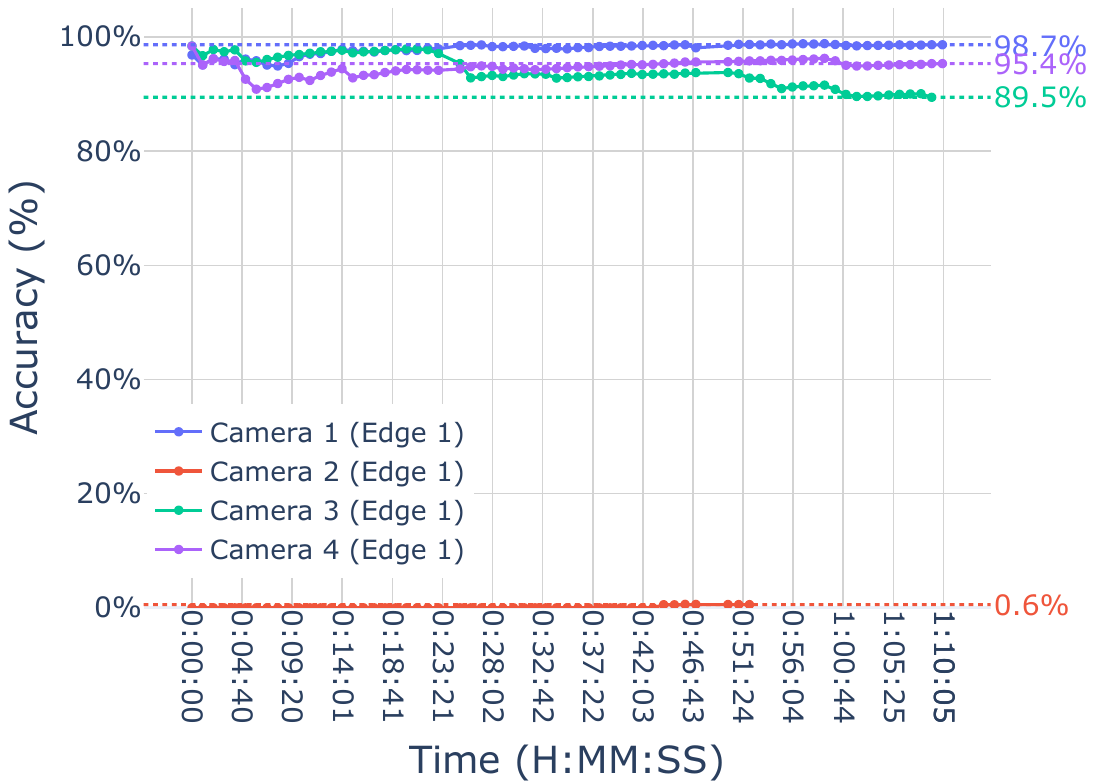}
  }
  \caption{Accuracy (Edge 1)}
  \label{fig:accur1}
\end{subfigure}

\vspace{1.0ex}

\begin{subfigure}{0.42\linewidth}
  \centering
  \raisebox{0pt}[\dimexpr\height][\depth]{
    \includegraphics[width=\linewidth]{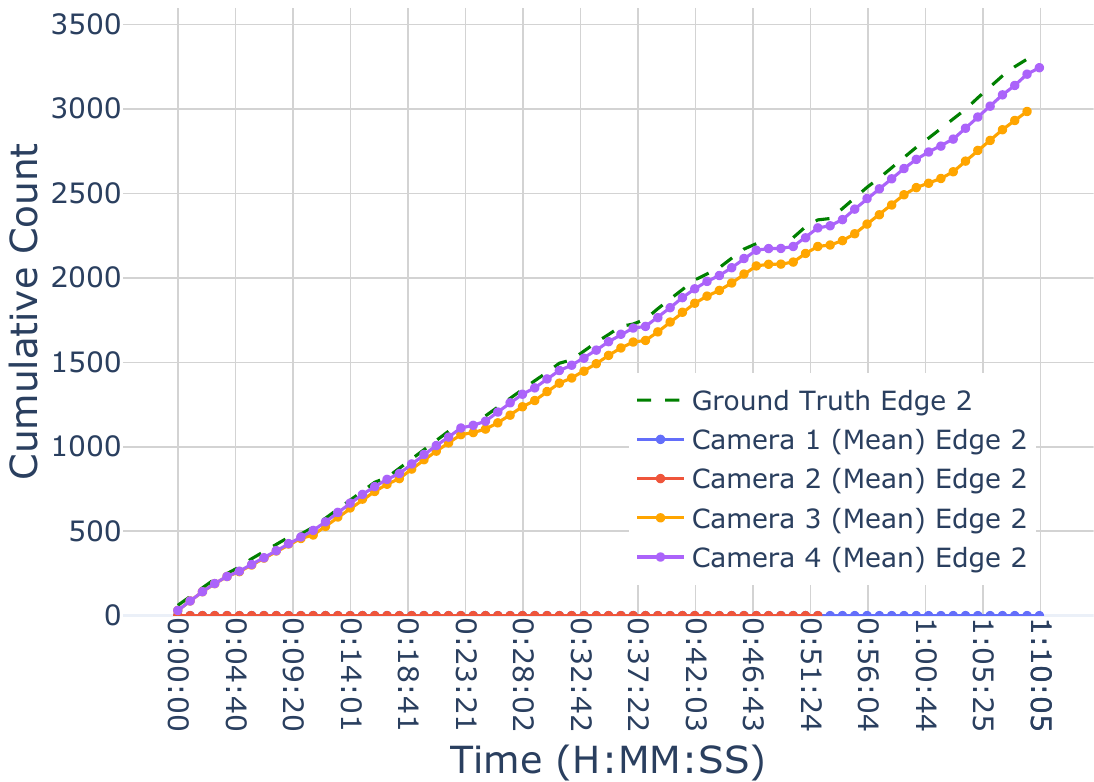}
  }
  \caption{Cumulative Count (Edge 2)}
  \label{fig:cumul2}
\end{subfigure}
\hspace{0.013\linewidth}
\begin{subfigure}{0.43\linewidth}
  \centering
  \raisebox{-2.2pt}[\dimexpr\height][\depth]{
    \includegraphics[width=\linewidth]{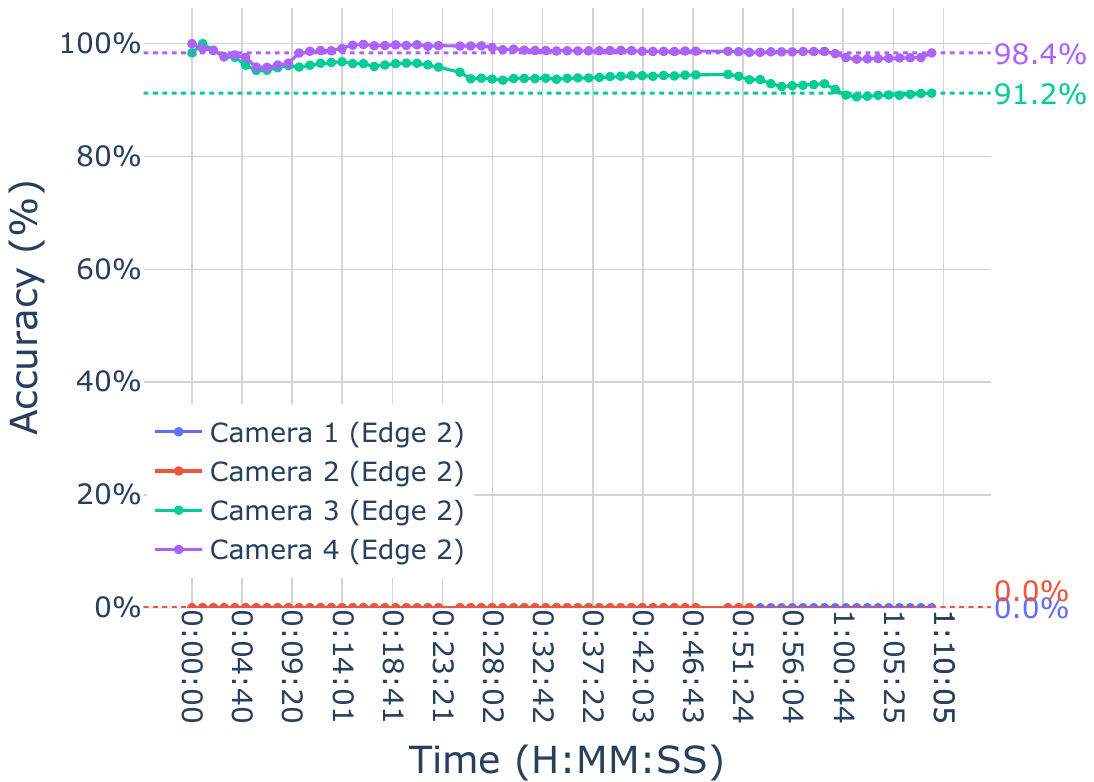}
  }
  \caption{Accuracy (Edge 2)}
  \label{fig:accur2}
\end{subfigure}

\caption{Performance of PerfCam's detection pipeline on two edges. Subplots (a) and (c) illustrate the close match between detected and ground truth product counts over time, while (b) and (d) highlight consistently high per-frame accuracy.}
\label{fig:od_results}
\end{figure*}

\subsubsection{Dataset and Extended Training Details}
\label{sec:dataset_training_details}

We used Roboflow~\cite{roboflow_universe_label_assist} to label a total of 29 frames collected from the four cameras (8 from Camera~3, 5 from Camera~2, 9 from Camera~1, and 7 from Camera~4).

From these frames, we performed an automated data preprocessing step that included:
\begin{itemize}
    \item Automatic orientation adjustment,
    \item Resizing each frame to $640 \times 640$ pixels, and 
    \item Converting the frames to grayscale.
\end{itemize}
In addition to preprocessing, we applied various data augmentation techniques to improve model robustness and reduce overfitting:
\begin{itemize}
    \item Flipping horizontally and vertically,
    \item Shearing (±10° horizontally, ±10° vertically),
    \item Adjusting saturation (between -25\% and +25\%),
    \item Introducing blur (up to 1.3\,pixels), and
    \item Adding noise (up to 1.13\% of the total pixels).
\end{itemize}
After augmentation, the dataset was expanded to 512 images in total, which were subsequently split into:
\begin{itemize}
    \item 440 images for the training set,
    \item 32 images for the validation set, and
    \item 40 images for the test set.
\end{itemize}

In total, after augmentations and splitting, 440 images were used to train our YOLO-based model. Once training was completed, we evaluated the model on the 32 validation images and 40 final test images. 

Following standard practices, we tracked both training and validation metrics over the course of training. Figure~\ref{fig:results_training} shows the training progress for the bounding box loss, classification loss, distribution focal loss, precision, recall, and the overlap between predicted bounding box output and the ground truth, also known as Intersection over Union (IoU).

To further validate the predicted bounding boxes, we performed a qualitative assessment on a separate validation set. Figure~\ref{fig:validation_images} shows an example batch of frames, with ground truth annotations in blue and the PerfCam’s predictions in orange with the detection confidence indicted next to each prediction. The strong overlap between predicted and labeled bounding boxes demonstrates the effectiveness of the approach in delineating individual products under typical conveyor conditions. However, there are cases where PerfCam fails to detect certain products or incorrectly detects objects, which suggests that additional training data and more thorough labeling could further improve its performance.

\begin{figure*}[htbp]
\centering

\begin{subfigure}[t]{0.32\linewidth}
  \centering
  \raisebox{0pt}[\dimexpr\height][\depth]{
    \includegraphics[width=\linewidth]{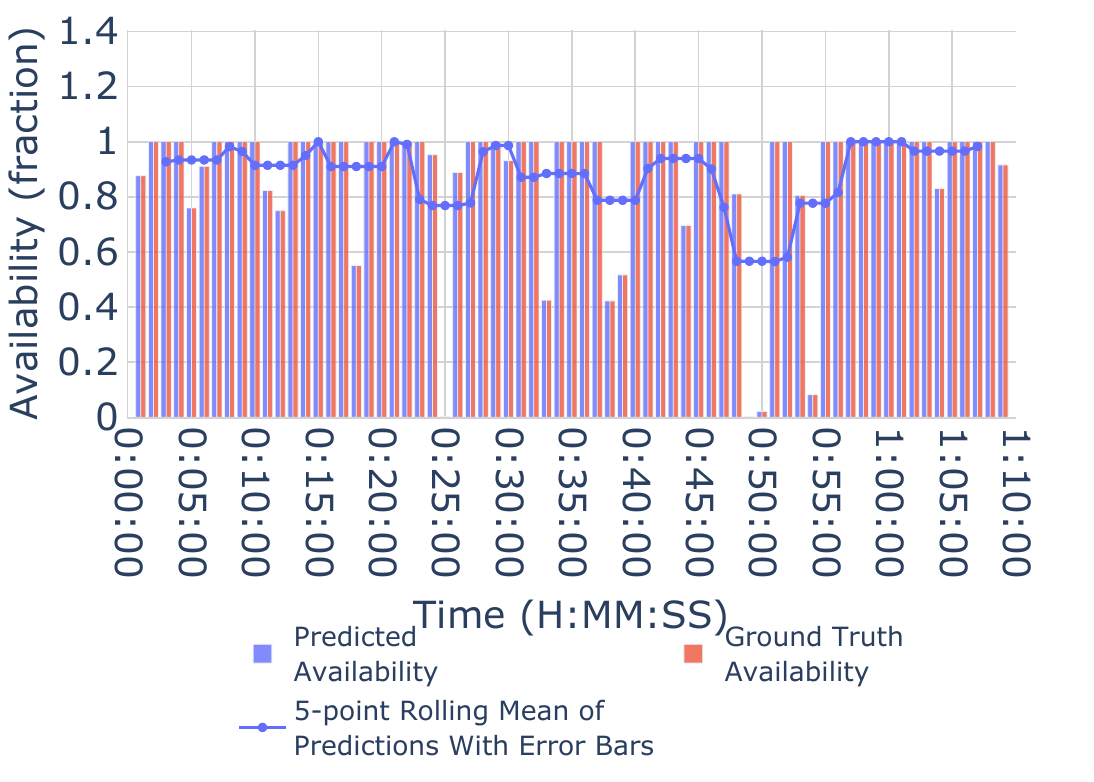}
  }
  \caption{Availability over time}
  \label{fig:avail}
\end{subfigure}
\hspace{0.001\linewidth}
\begin{subfigure}[t]{0.32\linewidth}
  \centering
  \raisebox{0pt}[\dimexpr\height][\depth]{
    \includegraphics[width=\linewidth]{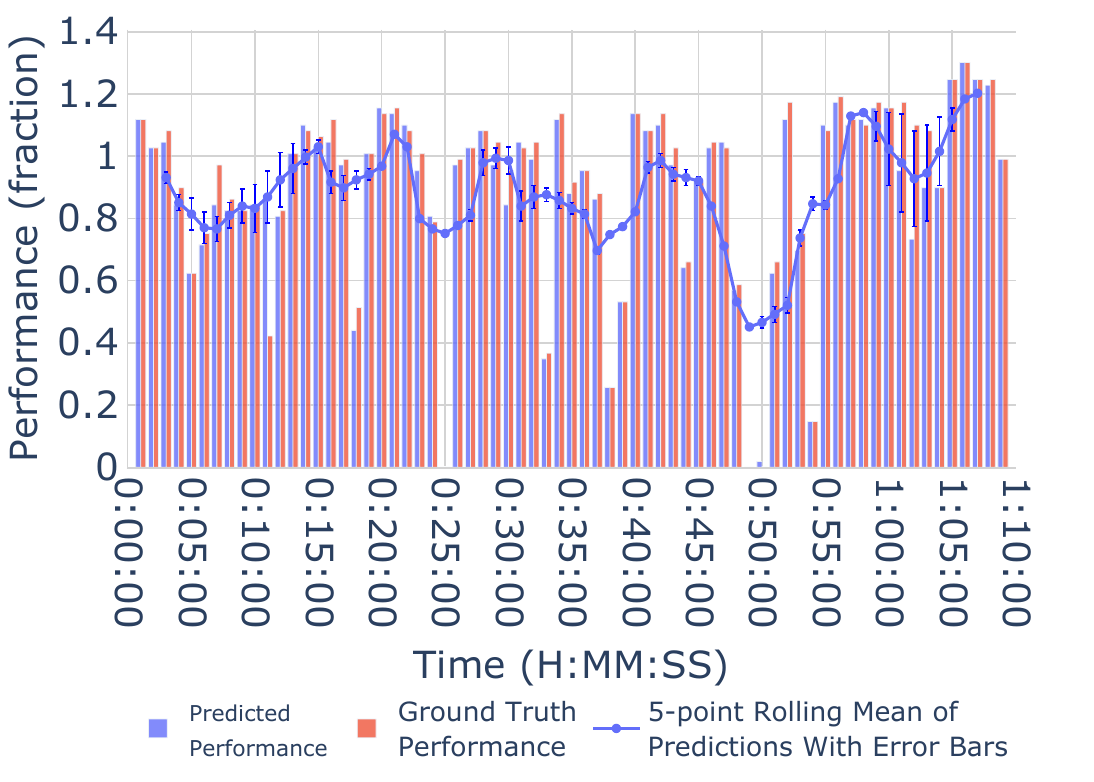}
  }
  \caption{Performance over time}
  \label{fig:perf}
\end{subfigure}
\hspace{0.001\linewidth}
\begin{subfigure}[t]{0.32\linewidth}
  \centering
  \raisebox{0pt}[\dimexpr\height][\depth]{
    \includegraphics[width=\linewidth]{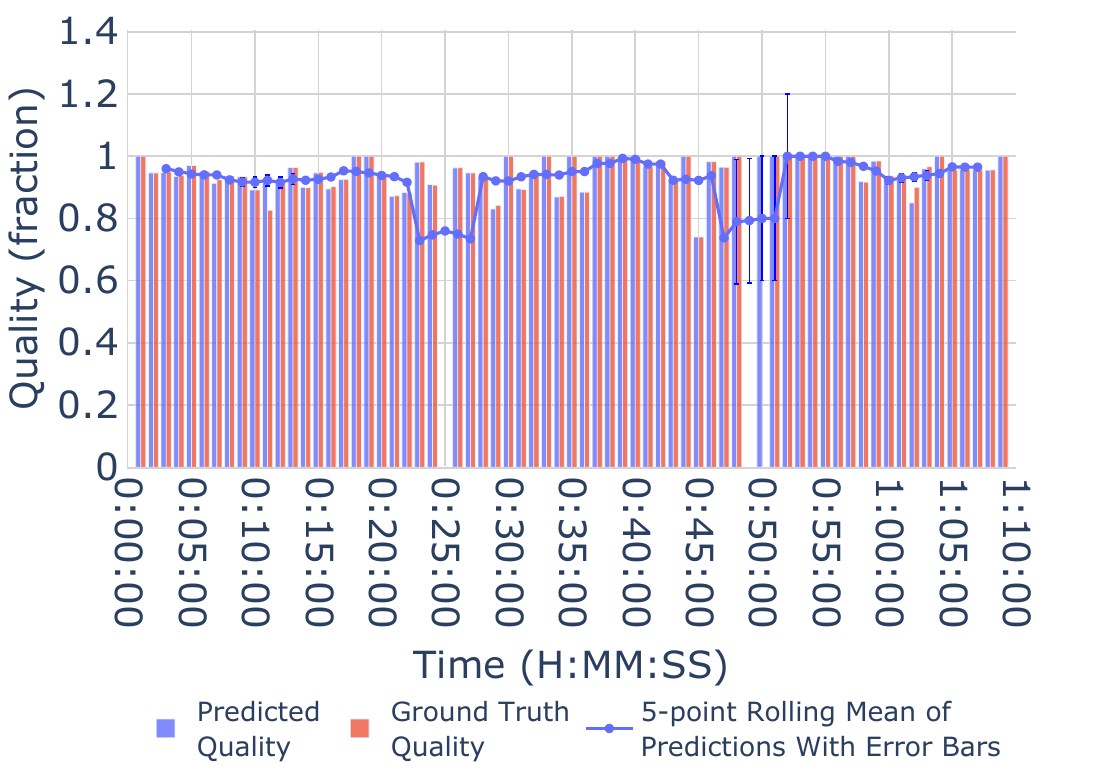}
  }
  \caption{Quality over time}
  \label{fig:qual}
\end{subfigure}

\vspace{1.0ex}

\begin{subfigure}[t]{0.32\linewidth}
  \centering
  \raisebox{0pt}[\dimexpr\height][\depth]{
    \includegraphics[width=\linewidth]{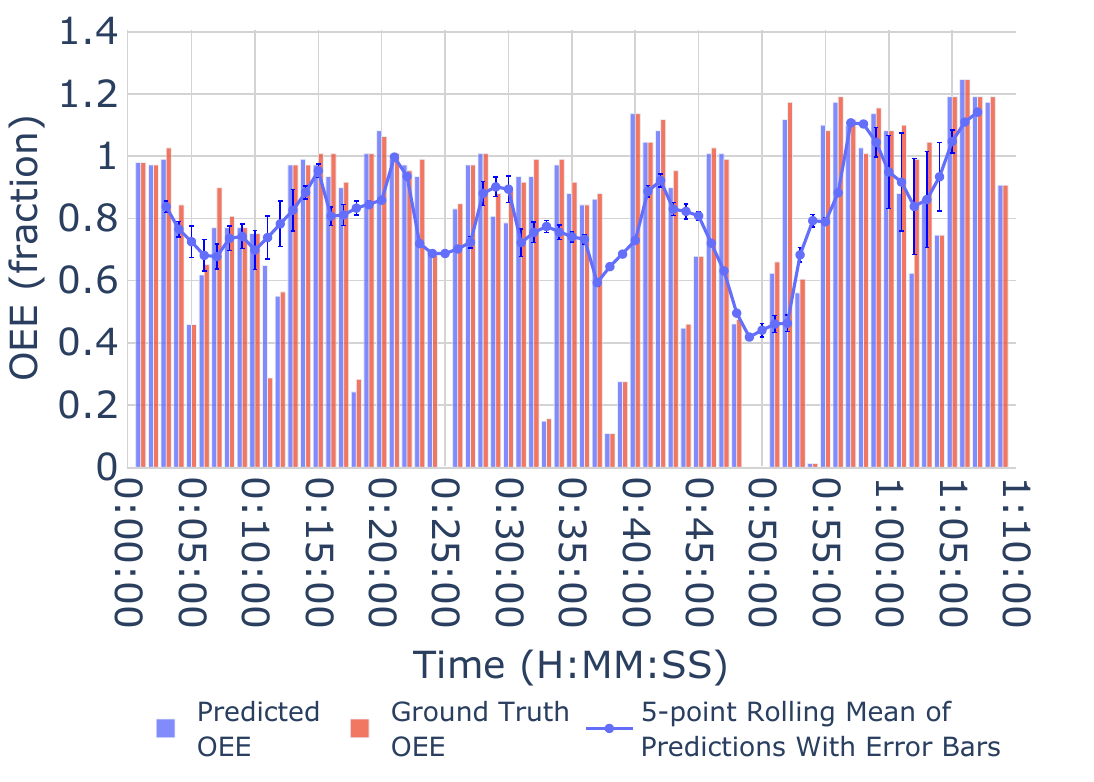}
  }
  \caption{OEE over time}
  \label{fig:oee}
\end{subfigure}
\hspace{0.001\linewidth}
\begin{subfigure}[t]{0.32\linewidth}
  \centering
  \raisebox{0pt}[\dimexpr\height][\depth]{
    \includegraphics[width=\linewidth]{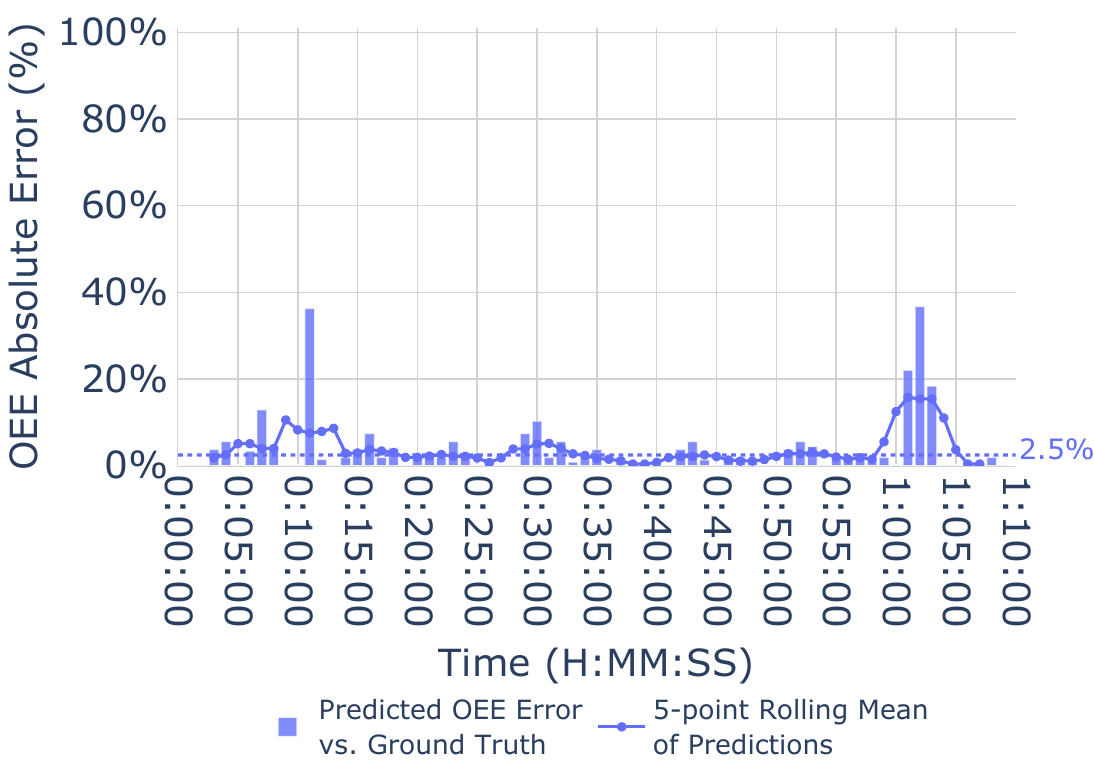}
  }
  \caption{Error of OEE predictions over time}
  \label{fig:oeeerr}
\end{subfigure}

\caption{Comparative OEE plots. Representing the availability, performance, quality, OEE, of predictions and ground truth (a-d) as well as the absolute error between PerfCam's measurements and ground truth over time (e).}
\label{fig:oee_stuff}
\end{figure*}

\subsubsection{Evaluation on the Test Set}

\label{sec:obj_detect_track}

To rigorously assess detection performance, we conducted parallel experiments on two edges, referred to as \emph{Edge 1} (connects Node 1 and Node 2) and \emph{Edge 2} (connects Node 2 and Node 3). In both setups, the pipeline continuously received frames from the surrounding cameras, analyzed them, and produced a stream of per-frame bounding boxes (in this experiment, this was done after collecting all the footage). We then compared these predictions against a ground truth dataset annotated by human experts. Figure~\ref{fig:od_results} summarizes two key metrics for each edge:

\begin{itemize}
    \item \textbf{Cumulative Count}: The cumulative number of products detected over time, compared with the ground truth count. 
    \item \textbf{Accuracy}: The percentage of correctly identified products per frame, measured at each time step.
\end{itemize}

\vspace{0.2em}
\noindent
\textbf{Cumulative Count vs. Ground Truth.}
Figures~\ref{fig:cumul1} and \ref{fig:cumul2} show the cumulative counts for Edge~1 and Edge~2, respectively. In both cases, the detection curve closely follows the ground truth line, suggesting only minor false negatives (i.e., products that occasionally went undetected) and a small average error of around 2--3\%. To reduce noise in the data and enhance the clarity of trends over time, the mean was calculated every minute.  Such near-alignment with ground truth indicates that the pipeline can reliably track item flow, which is crucial for calculating throughput and other downstream KPIs.

\vspace{0.2em}
\noindent

\textbf{Accuracy Over Time.} Before calculating the accuracy, we realized that variable frame rates in the recorded videos significantly affect the prediction results. Therefore, we transcode the videos and set the frame rates to a constant 30 fps before performing the accuracy calculations.

Figures~\ref{fig:accur1} and \ref{fig:accur2} present the frame-level detection accuracy for Edge~1 and Edge~2. Despite their differing hardware and computational capabilities, Edge 1 achieved a maximum accuracy of 98.7\%, while Edge 2 achieved a maximum accuracy of 98.4\%. This underscores the robustness of the pipeline in handling varying viewpoints, partial occlusions, and moderate lighting changes. Our camera with a low frame rate—identified as Camera 2—was unable to achieve correct tracking and resulted in extremely low accuracy (\textless{}1\%), highlighting the critical need for a higher recording frame rate to improve performance of tracking.

Additionally, Camera 1 exhibited less than 1\% accuracy on Edge~2 due to occlusion and the inability to maintain a clear view of Edge~2. To address this issue, Cameras 3 and 4 will be utilized as alternative viewpoints. In contrast, Cameras 1 and 4 maintained near-perfect visibility of the edges they are intended to monitor, ensuring higher accuracy in detection on corresponding edges.

Furthermore, gaps observed in parts of the accuracy plots are attributable to instances where the production line was stopped, as illustrated in the Availability plot in Figure~\ref{fig:avail_calc}. During these stoppages, no predictions were made because the tracking process was inactive, resulting in periods where accuracy could not be calculated. These interruptions manifest as gaps in the accuracy data, reflecting the temporary halt in the tracking pipeline.

Overall, these results confirm that our detection module, when integrated with the multi-camera setup and 3D reconstruction, can robustly localize and track products on the conveyor line.

\subsection{OEE and KPI Calculation}
\label{subsec:oee_kpi_calc}

To evaluate how well PerfCam measures standard production metrics, we compute four main KPIs: \emph{performance}, \emph{quality}, \emph{availability}, and \emph{OEE}. Figure~\ref{fig:oee_stuff} shows subplots (a)--(e) illustrate the progression of each KPI over time, from minute 1 to 69 together with a 5-point rolling mean, highlighting the interplay between availability, performance, and quality in determining the final OEE. These metrics capture how effectively the system is running, where any discrepancies from ground truth measurements inform how to improve the object detection pipeline.

\begin{itemize}
    \item \textbf{Throughput \& Performance:} We estimate throughput by counting how many products pass a designated reference line per minute. The \emph{performance} factor is then derived by comparing the empirical throughput to the ideal cycle time ($\tau_{\text{ideal}}=1.1\,\mathrm{s}$) and the time actually spent running (as shown in Figure~\ref{fig:perf}).
    \item \textbf{Quality:} We track how often a product is removed by the operator at \textbf{Node 2} (our manual ``QA station''), dividing the number of ``good'' products by the total number of products (as shown in Figure~\ref{fig:qual}). 
    \item \textbf{Availability:} In this experiment, we obtain availability purely from ground truth conveyor vibration data (see Section~\ref{sec:experiments} and Figure~\ref{fig:sensor_plots}), so there is no difference in comparison to a PerfCam-based availability prediction (as shown in Figure~\ref{fig:avail}). 
    \item \textbf{OEE:} This composite metric is computed by multiplying availability ($A$), performance ($P$), and quality ($Q$) (as shown in Figure~\ref{fig:oee}).
\end{itemize}

Figure~\ref{fig:oeeerr} presents the error between the OEE values calculated using PerfCam and those obtained from the ground truth, with the dotted line indicating the average error of 2.5\%. These results highlight how under- or overestimates of both throughput and defect rates can lead to deviations from the ground truth. In particular, from minutes~60--63, we tested a sub-scenario where packages were placed very close together on the conveyor—an arrangement not represented in the training data. As a result, multiple boxes merged into single detections, resulting in an undercount of the throughput and a visible drop in predicted OEE. Additionally, under this sub-scenario, the calculated OEE can exceed 1. This is because the production rate may temporarily appear higher than the ideal cycle time when products are extremely close together, which inflates the performance metric. Such occurrences, though rare in real production environments, highlight the importance of robust detection and tracking. Enhancing the training dataset with these challenging cases will improve the model's ability to accurately detect and track items under similar conditions, thereby reducing error rates and ensuring more reliable KPI calculations.

\vspace{0.5em}
\noindent

\section{Discussion on Challenges, Limitations, and Other Use Cases}
\label{sec:discussion}

In this section, we address the challenges and limitations encountered in the development and deployment of PerfCam as a framework, and explore potential extensions of it to other applications. Understanding these aspects is crucial for refining the system and expanding its applicability in diverse industrial and non-industrial contexts.

\subsection{Challenges}

Implementing PerfCam in industrial environments presents several challenges that need careful consideration. One significant challenge is the reliance on the quality of visual data and camera placement. Industrial settings often have space constraints, obstructed views, and safety and privacy regulations that limit optimal camera positioning. Insufficient camera coverage may create blind spots, reducing the effectiveness of 3D reconstruction and object detection. Moreover, environmental factors such as variable lighting conditions, reflections from metallic surfaces, and the presence of dust or fumes can degrade image quality. This can be mitigated through the strategic deployment of PerfCam in production lines, prioritizing high-value areas with minimal visual noise.

Another challenge lies in the computational demands of processing high-resolution images for real-time 3D reconstruction and object detection. Techniques like 3D Gaussian Splatting and advanced deep learning models require substantial processing power and memory. Deploying PerfCam on devices with limited computational resources may lead to increased latency and reduced performance, affecting the system's ability to provide timely insights. Optimizing algorithms for efficiency and exploring hardware acceleration options are essential to address this challenge.

Scalability is a key consideration when deploying the system in real-world applications. One of the main challenges is the significant number of cameras currently required to achieve complete coverage of the production line. This requirement can lead to increased complexity, higher costs, and greater maintenance demands, which may limit the solution's practicality in large-scale industrial settings. To address this issue, future research could focus on strategies to reduce camera dependency, such as optimizing camera placement, developing advanced stitching techniques, or utilizing predictive AI models to reconstruct data with fewer input streams. Furthermore, exploring alternative solutions like incorporating robots, drones, or telepresence technologies could also help mitigate these challenges. These strategies aim to improve the scalability of the system while maintaining its accuracy and overall performance.

Integrating PerfCam with existing industrial systems also poses challenges. Manufacturing environments often utilize a variety of legacy systems and proprietary protocols. Ensuring seamless communication and data exchange between PerfCam and these systems requires the development of compatible interfaces and adherence to industry communication standards. Additionally, deploying new technologies in a production environment must be done without disrupting ongoing operations, necessitating careful planning and possibly incremental integration strategies. Nevertheless, such challenges are a standard aspect of introducing new technologies and are not unique to PerfCam. With the appropriate methodologies and adherence to best practices, these issues can be effectively addressed.

\subsection{Limitations}

While PerfCam offers a flexible and cost-effective solution for monitoring and KPI extraction, there are inherent limitations to its approach. The system's performance is, like any other visual models, highly dependent on the availability and quality of visual data. In situations where cameras cannot be installed optimally or where the environment poses challenges like poor lighting or occlusions, the accuracy of the system may decrease. For example, in areas with intense glare or shadows, the object detection algorithms might fail to recognize and track items accurately.

Another limitation is the system's focus on visual data, which may not capture all relevant parameters of a production process. Certain metrics, such as internal temperatures, pressures, vibrations, or chemical compositions, cannot be assessed through visual means alone. Relying solely on visual data may provide an incomplete picture of the operational state, potentially overlooking critical factors that influence efficiency and safety.

Data privacy and security are also important considerations. The use of cameras to capture detailed visual information can raise concerns about exposing sensitive processes or proprietary information. Ensuring that camera footage and sensory information is securely stored and transmitted, and that access is controlled and monitored, is crucial to protect confidentiality, data privacy, and adhere to regulatory requirements.

\subsection{Other Use Cases}

The technologies and methodologies developed in PerfCam have applications beyond industrial production lines. In example, smart city initiatives could utilize the system for traffic management, public safety, and infrastructure monitoring. By leveraging digital twin technology, similar to what PerfCam offers, to deliver detailed, real-time insights on urban environments, city planners and administrators can make data-driven decisions that optimize services and enhance residents’ quality of life.

In summary, while there are challenges and limitations involved in introducing PerfCam to industrial settings, the underlying computer vision based digital twinning technique offer considerable potential for a wide range of applications. Addressing the identified challenges through technical innovations and adaptations can extend the benefits of the platform to various industries and use cases.

\section{Conclusion, Limitations, and Future Works}
\label{sec:conclusion}

In this paper, we introduced PerfCam, an open-source PoC digital twinning framework that leverages 3D Gaussian Splatting, sensory data, and real-time object detection for digital twinning and KPI extraction in industrial production lines. Our experiments in a semi-production environment demonstrated that PerfCam can reconstruct 3D scenes, detect and track objects, and extract essential KPIs such as throughput, conveyor belt speed, and OEE using visual data.

\textbf{Contributions.} PerfCam integrates 3D Gaussian Splatting techniques, computer vision models, and sensor fusion together with annotated visualizations to construct digital twins that can visualize some of the production KPIs. By leveraging existing camera infrastructure, as well as incorporating additional cameras or sensors when necessary, PerfCam offers a cost-effective and scalable solution suitable for various industrial settings. Through real-world experimentation on an industrial test line, we demonstrate the system’s capabilities for real-time 3D reconstruction, object detection, and automated KPI extraction (e.g., OEE, throughput, and conveyor speed). These features collectively aid in identifying bottlenecks, enhancing production oversight, and guiding process optimization. Furthermore, by openly publishing both our framework and dataset, we aim to foster ongoing research and encourage broader adoption of camera-based digital twins in modern manufacturing environments.

\textbf{Limitations.} Despite the promising results during our evaluation phase, there are several limitations to our approach. First, the accuracy of the 3D reconstruction and object detection is dependent on the quality and quantity of input images. In environments with limited camera placements or challenging lighting conditions, the performance may degrade. Second, the computational requirements for real-time processing can be substantial, especially for high-resolution images and complex scenes, which may limit deployment on devices with constrained resources. Third, while our system handles occlusions and lighting variations to some extent, extreme conditions can still pose challenges to the object detection and tracking modules.

\textbf{Future Work.} Future work includes optimizing the system for better performance on limited hardware, adding more predictive capabilities, enhancing robustness to extreme environmental conditions, reducing the need for excessive number of cameras, possibility to simulate different scenarios, and extending the platform to support additional KPIs and analytics. As outlined throughout the paper, PerfCam is currently an open-source PoC, and many of its components, steps, and modules have yet to be integrated into a unified platform to qualify as a stand-alone and fully functional prototype such that a potentially production ready solution can be developed. Therefore, a potential future work could be for researchers and engineers, in both academia and industry, to develop such prototype while mitigating the aforementioned challenges. There is also a potential to leverage general transformer-based models (such as those provided by Large Language Models) for broad assessments, while training more specialized models—such as advanced generative frameworks for simulating rare production line scenarios or convolutional neural networks for real-time defect detection—to provide targeted recommendations based on visual data. In particular, advanced generative frameworks can produce realistic synthetic data—including unusual or rare conditions—that enrich both the training pipeline and high-level system evaluations. By simulating these edge cases, the system is stress-tested against potential failures, leading to improved robustness and adaptability across the entire production line. An extensive evaluations in fully operational production lines would also further validate and improve the system's limitations, and enable cost comparisons.

\section*{Acknowledgment}
The authors gratefully acknowledge the support of Digital Futures at KTH Royal Institute of Technology and AstraZeneca for funding and enabling this research through the Industrial Postdoc Project \textit{SMART: Smart Predictive Maintenance for the Pharmaceutical Industry}, grant no. KTH-RPROJ-0146472. The computations and data-handling were partially enabled by resources at Alvis provided by the National Academic Infrastructure for Supercomputing in Sweden (NAISS), partially funded by the Swedish Research Council through grant agreement no. 2022-06725. The authors extend their heartfelt gratitude to Sam Bradley, David Gray Lassiter, Tianzhi Li, and Lihui Wang for their invaluable support and feedback throughout the course of this work.

\bibliographystyle{IEEEtran}
\bibliography{IEEEabrv,main}

\end{document}